%% file: main.tex
\documentclass{article}

\usepackage[table,xcdraw]{xcolor}

\usepackage[T1]{fontenc}
\usepackage[utf8]{inputenc}

\RequirePackage{silence}
\usepackage[inline]{enumitem}
\usepackage{siunitx}
\usepackage[hyphens]{url}  %
\usepackage{stringstrings}  %

\usepackage{natbib}
\setcitestyle{authoryear,round,citesep={;},aysep={,},yysep={;}}

\usepackage{adjustbox}
\usepackage{chngcntr}
\usepackage{microtype}
\usepackage{graphicx}
\usepackage{subcaption}
\usepackage{booktabs} %
\usepackage{multirow}
\usepackage{multicol}
\usepackage{pgfplots}
\usepackage[section]{placeins}
\usepackage{capt-of}
\usepackage{svg}
\usepackage{tikz-cd}
\usepackage{tikz}
\usetikzlibrary{shapes.geometric}
\usetikzlibrary{positioning}

\usepackage{amsmath}
\usepackage{amsfonts}
\usepackage{amssymb}
\usepackage{amsthm}
\usepackage{bbm}
\usepackage{thmtools}
\usepackage{mathtools}

\DeclarePairedDelimiter\floor{\lfloor}{\rfloor}

\newcommand{\envname}[1]{\texttt{#1}}
\newcommand{\algoname}[1]{\texttt{#1}}

\newcommand{\taskheading}[2]{\envname{#1}: #2}

\tikzstyle{circ} = [circle, text centered, draw=black]

\usepackage{hyperref}

\usepackage{arxiv}
\usepackage{abbreviations}

\def\isaccepted{}
\ifdefined\isaccepted
\newcommand{\projectsource}{\url{https://github.com/HumanCompatibleAI/seals}}
\newcommand{\algorithmsource}{\url{https://github.com/HumanCompatibleAI/derail}}
\else
\newcommand{\projectsource}{\textit{--- removed for double-blind ---}}
\newcommand{\algorithmsource}{\textit{--- removed for double-blind ---}}
\fi

\pgfplotsset{compat=1.16}
\begin{document}

\setlist[itemize]{itemsep=2pt,parsep=0pt,topsep=2pt}  %

\title{DERAIL: Diagnostic Environments for Reward And Imitation Learning}

\ifdefined\isaccepted
\author{Pedro Freire\thanks{Work partially conducted during an internship at UC Berkeley.} \\
  École Polytechnique \\
  \texttt{pedrofreirex@gmail.com}
	\And
	Adam Gleave \\
	UC Berkeley \\
    \texttt{gleave@berkeley.edu}
	\And
	Sam Toyer \\
	UC Berkeley \\
    \texttt{sdt@berkeley.edu}
	\And
	Stuart Russell \\
	UC Berkeley \\
    \texttt{russell@berkeley.edu}
}
\else
\author{Anonymous Authors}
\fi

\vskip 0.3in

\maketitle

\begin{abstract}
The objective of many real-world tasks is complex and difficult to procedurally specify.
This makes it necessary to use reward or imitation learning algorithms to infer a reward or policy directly from human data.
Existing benchmarks for these algorithms focus on realism, testing in complex environments.
Unfortunately, these benchmarks are slow, unreliable and cannot isolate failures.
As a complementary approach, we develop a suite of simple diagnostic tasks that test individual facets of algorithm performance in isolation.
We evaluate a range of common reward and imitation learning algorithms on our tasks.
Our results confirm that algorithm performance is highly sensitive to implementation details.
Moreover, in a case-study into a popular preference-based reward learning implementation, we illustrate how the suite can pinpoint design flaws and rapidly evaluate candidate solutions.
The environments are available at \projectsource{}.
\end{abstract}

\input{intro}
\input{desiderata}
\input{tasks}
\input{experiments}
\input{case_study}
\input{discussion}

\ifdefined\isaccepted
\section*{Acknowledgements}
We would like to thank Rohin Shah and Andrew Critch for feedback during the initial stages of this project, and Scott Emmons, Cody Wild, Lawrence Chan, Daniel Filan and Michael Dennis for feedback on earlier drafts of the paper.
\fi

\bibliographystyle{plainnat}
\bibliography{refs}

\clearpage
\appendix
\counterwithin{table}{section}
\counterwithin{figure}{section}

\input{task_specs}
\input{experiments_full}

\end{document}

%% file: intro.tex
\section{Introduction}

Reinforcement learning (RL) optimizes a fixed reward function specified by the designer.
This works well in artificial domains with well-specified reward functions such as games~\citep{silver:2016,vinyals:2019,openai:2019}.
However, in many real-world tasks the agent must interact with users who have complex and heterogeneous preferences.
We would like the AI system to satisfy users' preferences, but the designer cannot perfectly anticipate users' desires, let alone procedurally specify them.
This challenge has led to a proliferation of methods seeking to learn a reward function from user data~\citep{ng:2000,ziebart:2008,christiano:2017,fu:2018,cabi:2019}, or imitate demonstrations~\citep{ross:2011,ho:2016,reddy:2020}.
Collectively, we say algorithms that learn a reward or policy from human data are \emph{Learning from Humans (LfH)}.

LfH algorithms are primarily evaluated empirically, making benchmarks critical to progress in the field.
Historically, evaluation has used RL benchmark suites.
In recognition of important differences between RL and LfH, recent work has developed imitation learning benchmarks in complex simulated robotics environments with visual observations~\citep{memmesheimer:2019,james:2020}.

In this paper, we develop a complementary approach using simple diagnostic environments that test individual aspects of LfH performance in isolation.
Similar diagnostic tasks have been applied fruitfully to RL~\citep{osband:2020}, and diagnostic datasets have long been popular in natural language processing~\citep{johnson:2017,sinha:2019,kottur:2019,liu:2019,wang:2019}.
Diagnostic tasks are analogous to unit-tests: while less realistic than end-to-end tests, they have the benefit of being fast, reliable and able to isolate failures~\citep{myers2011art,wacker:2015}.
Isolating failure is particularly important in machine learning, where small implementation details may have major effects on the results~\citep{islam2017reproducibility}.

This paper contributes the first suite of diagnostic environments designed for LfH algorithms.
We evaluate a range of LfH algorithms on these tasks.
Our results in section~\ref{sec:experiments} show that, like deep RL~\citep{henderson2018deep,engstrom:2020}, imitation learning is very sensitive to implementation details.
Moreover, the diagnostic tasks isolate particular implementation differences that affect performance, such as positive or negative bias in the discriminator.
Additionally, our results suggest that a widely-used preference-based reward learning algorithm~\citep{christiano:2017} suffers from limited exploration.
In section~\ref{sec:case-study}, we propose and evaluate several possible improvements using our suite, illustrating how it supports rapid prototyping of algorithmic refinements.

\begin{figure}
    \begin{minipage}[b]{0.45\textwidth}
      \vspace{10pt}
      \centering
      \scalebox{1.2}{
      \input{figs/tasks/risky_path.tikz}
      }
      \caption{\envname{RiskyPath}: The agent can either take a long but sure path to the goal ($s_0 \to s_1 \to s_2$), or attempt to take a shortcut ($s_0 \to s_2)$, with the risk of receiving a low reward ($s_0 \to s_3$).}
      \label{fig:task:risky-path}
    \end{minipage}
    \hfill
    \begin{minipage}[b]{0.45\textwidth}
      \centering
      \begin{subfigure}{0.98\textwidth}
        \centering
        \input{figs/tasks/early_termination_pos.tikz}
      \end{subfigure}

      \begin{subfigure}{0.98\textwidth}
        \centering
        \input{figs/tasks/early_termination_neg.tikz}
      \end{subfigure}
      \caption{\earlyp{} (top) or \earlyn{} (bottom): The agent can either alternate between the first two states until the horizon ends, or end the episode early by moving to the terminal state (far right).}
      \label{fig:task:early_term_pos}
    \end{minipage}
\end{figure}

%% file: figs/tasks/risky_path.tikz
\begin{tikzpicture}[node distance = 0.5cm, thick]
    \node (0) [circ] {$s_0$};
    \node (1) [circ, above=0.25cm of 0, xshift=1cm]{$s_1$};
    \node (r) [draw=black, right=1cm of 0]{};
    \node (2) [circ, fill=green!10, right=1cm of r, yshift=0.7cm]{$s_2$};
    \node (3) [circ, fill=red!70, below=0.3cm of 2]{$s_3$};

    \node (rew-2) [right=of 2] {$r = 1.0$};
    \node (rew-3) [right=of 3] {$r = -100.0$};

    \draw[->] (0) to [bend left] (1);
    \draw[->] (1) to [bend left] (2);
    \draw[-] (0) to (r);
    \draw[->] (r) to [bend left] (2);
    \draw[->] (r) to [bend right] (3);

    \draw[->] (r) edge[bend left] node[pos=0.5,rotate=40,above]{$50\%$} (2);
    \draw[->] (r) edge[bend right] node[pos=0.5,rotate=-40,below]{$50\%$} (3);

    \draw[->] (1) edge [loop above] (1);
    \draw[->] (2) edge [loop right] (2);
    \draw[->] (3) edge [loop right] (3);
\end{tikzpicture}

%% file: figs/tasks/early_termination_pos.tikz
\begin{tikzpicture}[node distance = 1.0cm, thick, minimum size=0.8cm]
  % \node (1) [circ] {$s_0$};

  \node (1) [circ, inner sep=0cm] {\includegraphics[width=0.55cm]{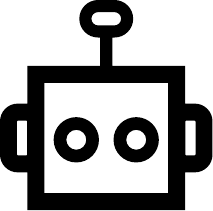}};
  \node (2) [circ, right=of 1] {};
  \node (3) [circ, right=of 2, inner sep=0cm] {\includegraphics[width=0.45cm]{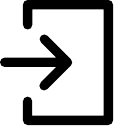}};

  \node (r1) [above=of 1, yshift=-0.8cm]{r = 1.0};
  \node (r2) [above=of 2, yshift=-0.8cm]{r = 1.0};
  \node (r3) [above=of 3, yshift=-0.8cm]{r = 1.0};

  \draw [->] (1) to [bend left] (2);
  \draw [->] (2) to [bend left] (1);
  \draw [->] (2) -- (3);
\end{tikzpicture}

%% file: figs/tasks/early_termination_neg.tikz
\begin{tikzpicture}[node distance = 1.0cm, thick, minimum size=0.8cm]
  \node (1) [circ, inner sep=0cm] {\includegraphics[width=0.55cm]{figs/icons/noun_robot_head_771010.pdf}};
  \node (2) [circ, right=of 1] {};
  \node (3) [circ, right=of 2, inner sep=0cm] {\includegraphics[width=0.45cm]{figs/icons/noun_Exit_2494210.pdf}};

  \node (r1) [above=of 1, yshift=-0.8cm]{r = -1.0};
  \node (r2) [above=of 2, yshift=-0.8cm]{r = -1.0};
  \node (r3) [above=of 3, yshift=-0.8cm]{r = -1.0};

  \draw [->] (1) to [bend left] (2);
  \draw [->] (2) to [bend left] (1);
  \draw [->] (2) -- (3);
\end{tikzpicture}

%% file: desiderata.tex
\section{Designing Diagnostic Tasks}
\label{sec:desiderata}

In principle, LfH algorithms can be evaluated in any Markov decision process.
Designers of benchmark suites must reduce this large set of possibilities to a small set of tractable tasks that discriminate between algorithms.
We propose three key desiderata to guide the creation of diagnostic tasks.

\textbf{Isolation}.
Each task should test a single dimension of interest.
The dimension could be a capability, such as robustness to noise; or the absence of a common failure mode, such as episode termination bias~\cite{kostrikov2018discriminator}.
Keeping tests narrow ensures failures pinpoint areas where an algorithm requires improvement.
By contrast, an algorithm's performance on more general tasks has many confounders.

\textbf{Parsimony}.
Tasks should be as simple as possible.
This maintains compatibility with a broad range of algorithms.
Furthermore, it ensures the tests run quickly, enabling a more rapid development cycle and sufficient replicas to achieve low-variance results.
However, tasks may need to be computationally demanding in special cases, such as testing if an algorithm can scale to high-dimensional inputs.

\textbf{Coverage}.
The benchmark suite should test a broad range of capabilities.
This gives confidence that an algorithm passing all diagnostic tasks will perform well on general-purpose tasks.
For example, a benchmark suite might want to test categories as varied as exploration ability, the absence of common design flaws and bugs, and robustness to shifts in the transition dynamics.

%% file: tasks.tex
\section{Tasks}

\newcommand{\taskdivider}[1]{\subsubsection{#1}}

In this section, we outline a suite of tasks we have developed around the
guidelines from the previous section. Some of the tasks have
configuration parameters that allow the difficulty of the task to be adjusted.
A full specification of the tasks can be found in appendix~\ref{sec:appendix:tasks-specification}.

\subsection{Design Flaws and Implementation Bugs}

First, we describe tasks that check for fundamental issues in the design and implementation of algorithms.

\taskdivider{\taskheading{RiskyPath}{Stochastic Transitions}}
Many LfH algorithms are derived from Maximum Entropy Inverse Reinforcement Learning~\citep{ziebart:2008}, which models the demonstrator as producing trajectories with probability $p(\tau) \propto \exp R(\tau)$.
This model implies that a demonstrator can ``control'' the environment well
enough to follow any high-reward trajectory with high
probability~\citep{ziebart:2010:thesis}. However, in stochastic environments,
the agent cannot control the probability of each trajectory independently.
This misspecification may lead to poor behavior.

To demonstrate and test for this issue, we designed \risky{}, illustrated in
Figure~\ref{fig:task:risky-path}. The agent starts at $s_0$ and can reach the
goal $s_2$ (reward $1.0$) by either taking the \emph{safe} path $s_0 \to
s_1 \to s_2$, or taking a \emph{risky} action, which has equal chances of going
to either $s_3$ (reward $-100.0$) or $s_2$.
The \emph{safe} path has the highest expected return, but the \emph{risky} action sometimes reaches the goal $s_2$ in fewer timesteps, leading to higher best-case return.
Algorithms that fail to correctly handle stochastic dynamics may therefore wrongly believe the reward favors taking the \emph{risky} path. 

\taskdivider{\taskheading{EarlyTerm}{Early Termination}}
Many implementations of imitation learning algorithms incorrectly assign a value of zero to terminal states~\citep{kostrikov2018discriminator}.
Depending on the sign of the learned reward function in non-terminal states, this can either bias the agent to end episodes early or prolong them as long as possible.
This confounds evaluation as performance is spuriously high in tasks where the termination bias aligns with the task objective.
\citeauthor{kostrikov2018discriminator} demonstrate this behavior with a simple example, which we adapt here as the tasks \earlyp{} and \earlyn{}.

The environment is a 3-state MDP, in which the agent can either alternate between two initial states until reaching the time horizon, or they can move to a terminal state causing the episode to terminate early.
In \earlyp{}, the rewards are all $+1$, while in \earlyn{}, the rewards are all $-1$.
Algorithms that are biased towards early termination (e.g.\ because they assign a negative reward to all states) will do well on \earlyn{} and poorly on \earlyp{}.
Conversely, algorithms biased towards late termination will do well on \earlyp{} and poorly on \earlyn{}.

\subsection{Core Capabilities}

In this subsection, we consider tasks that focus on a core algorithmic capability for reward and imitation learning.

\begin{figure}
    \begin{minipage}[b]{0.45\textwidth}
      \begin{subfigure}{0.45\textwidth}
        \centering
        \input{figs/tasks/noisy_obs_states.tikz}
        \caption{Underlying state space}
      \end{subfigure}
      \begin{subfigure}{0.45\textwidth}
        \centering
        \scalebox{0.8}{
          \input{figs/tasks/noisy_obs_obvector.tikz}
        }
        \caption{Observation vector}
      \end{subfigure}
      \caption{\envname{NoisyObs}: Agent starts at a random corner and gets a
        reward for staying at the center of the grid. The observation consists of the agent's $x$-$y$ coordinates concatenated with Gaussian noise.}
        \label{fig:noisy-obs}
    \end{minipage}
    \hfill
    \begin{minipage}[b]{0.45\textwidth}
      \centering
      \scalebox{0.9}{
        \input{figs/tasks/branching.tikz}
      }
      \caption{\branch{}: Hard-exploration problem. Agent must follow a long
        path without taking any wrong actions; the agent only gets a non-zero
        reward by performing the optimal trajectory.}
      \label{fig:branching}
    \end{minipage}

\end{figure}

\taskdivider{\taskheading{NoisyObs}{Robust Learning}}
\noisyobs{}, illustrated in Figure~\ref{fig:noisy-obs}, tests for robustness to noise.
The agent randomly starts at the one of the corners of an $M \times M$ grid (default $M = 5$), and tries to reach and stay at the center.
The observation vector consists of the agent's $(x,y)$ coordinates in the first two elements, and $L$ ``distractor'' samples of Gaussian noise as the remaining elements (default $L=20$).
The challenge is to \emph{select} the relevant features in the observations, and not overfit to noise~\citep{guyon:2003}.

\taskdivider{\taskheading{Branching}{Exploration}}
We include the \envname{Branching} task to test LfH algorithms' exploration ability. The agent must traverse a specific path of length $L$ to reach a
final goal (default $L=10$), with $B$ choices at each step (default $B=2$). Making the wrong choice at any of the $L$ decision points leads to a dead end with zero reward.

\begin{figure}
    \begin{minipage}[b]{0.45\textwidth}
      \centering
      \scalebox{1.6}{
        \input{figs/tasks/parabola.tikz}
      }
      \caption{\parabola{}: A random parabola is sampled at the start of the
        episode; the agent moves horizontally at a constant speed, and must
        adjust its $y$ coordinate to match the curve.}
      \label{fig:parabola}
    \end{minipage}
    \hfill
    \begin{minipage}[b]{0.45\textwidth}
      \centering
      \scalebox{0.9}{
        \input{figs/tasks/largest_sum.tikz}
      }
      \vspace{10pt}
      \caption{\largest{}: Simple linear classification problem. The state is a
        high-dimensional vector and the goal is to output which half of the
        vector has the largest sum.}
      \label{fig:largest}
    \end{minipage}

\end{figure}

\taskdivider{\taskheading{Parabola}{Continuous Control}}
\envname{Parabola} tests algorithms' ability to learn in continuous action spaces, a challenge for $Q$-learning methods in particular.
The goal is to mimic the path of a parabola $p(x) = Ax^2 + Bx + C$, where $A$, $B$ and $C$ are constants sampled
uniformly from $[-1, 1]$ at the start of the episode.
The state at time $t$ is $s_t = (x_t, y_t, A, B, C)$.
Transitions are given by $x_{t+1} = x_t + dx$ (default $dx = 0.05$) and $y_{t+1} = y_t + a_t$.
The reward at each timestep is the negative squared error, $-\left(y_t-p(x_t)\right)^2$.

\taskdivider{\taskheading{LargestSum}{High Dimensionality}}
Many real-world tasks are high-dimensional.
\envname{LargestSum} evaluates how algorithms scale with increasing dimensionality.
It is a classification task with binary actions and uniformly sampled states $s \in [0, 1]^{2L}$ (default $L = 25$).
The agent is rewarded for taking action $1$ if the sum of the first half $x_{0:L}$ is greater than the sum of the second half $x_{L:2L}$, and otherwise is rewarded for taking action $0$.

\subsection{Ability to Generalize}

In complex real-world tasks, it is impossible for the learner to observe every state during training, and so some degree of generalization will be required at deployment.
These tasks simulate this challenge by having a (typically large) state space which is only partially explored at training.

\taskdivider{\taskheading{InitShift}{Distribution Shift}}
Many LfH algorithms learn from expert demonstrations.
This can be problematic when the environment the demonstrations were gathered in differs even slightly from the learner's environment.

To illustrate this problem, we introduce \initstate{}, a depth-2 full binary tree where the agent moves left or right until reaching a leaf.
The expert starts at the root $s_0$, whereas the learner starts at the left branch $s_1$ and so can only reach leaves $s_3$ and $s_4$.
Reward is only given at the leaves.
The expert always move to the highest reward leaf $s_6$, so any algorithm that relies on demonstrations will not know whether it is better to go to $s_3$ or $s_4$.
By contrast, feedback such as preference comparison can disambiguate this case.

\taskdivider{\taskheading{ProcGoal}{Procedural Generation}}
In this task, the agent starts at a random position in a large grid, and must navigate to a goal randomly placed in a neighborhood around the agent.
The observation is a 4-dimensional vector containing the $(x,y)$ coordinates of the agent and the goal. 
The reward at each timestep is the negative Manhattan distance between the two positions.
With a large enough grid, generalizing is necessary to achieve good performance, since most initial states will be unseen.

\taskdivider{\taskheading{Sort}{Algorithmic Complexity}}
In \envname{Sort}, the agent must sort a list of random numbers by swapping
elements. The initial state is a vector $x$ sampled uniformly from $[0,1]^n$
(default $n=4$), with actions $a = (i,j)$ swapping $x_i$ and $x_j$.
The reward is given according to the number of elements in the correct position.
To perform well, the learned policy must compare elements, otherwise it will not generalize to all possible randomly selected initial states.

\begin{figure}
\begin{minipage}[b]{0.30\textwidth}
  \centering
  \scalebox{1.0}{
    \input{figs/tasks/init_state_shift.tikz}
  }
  \caption{\initshift{}: Expert starts at root $s_0$, learner starts at $s_0^*$.
    Optimal expert demonstrations go to lower branch and so are uninformative about the uppper branch $s_0^*$.}
  \label{fig:initshift}
\end{minipage}
\hfill
\begin{minipage}[b]{0.30\textwidth}
  \centering
  \scalebox{0.9}{
    \input{figs/tasks/procedural_goal.tikz}
  }
  \caption{\procgoal{}: Agent (robot) and goal (green cell) are randomly positioned in a large
    grid. The agent sees only a small fraction of possible states during training.}
  \label{fig:procgoal}
\end{minipage}
\hfill
\begin{minipage}[b]{0.30\textwidth}
  \centering
  \scalebox{0.8}{
    \input{figs/tasks/sort.tikz}
  }
  \caption{\sort{}: The agent has to sort a list by swapping elements; to
    perform well, policies and rewards must learn to perform comparisons between elements.}
  \label{fig:sort}
\end{minipage}
\end{figure}

%% file: figs/tasks/noisy_obs_states.tikz
 \begin{tikzpicture}[node distance = 1.5cm, thick]% 
      \draw[step=0.8cm,color=gray] (0, 0) grid (2.4, 2.4);

      \node[color=gray, opacity=0.1] (A) at (0.4, 2.0) {\includegraphics[width=0.65cm]{figs/icons/noun_robot_head_771010.pdf}};
      \node[color=gray, opacity=1.0] (A) at (0.4, 1.2) {\includegraphics[width=0.65cm]{figs/icons/noun_robot_head_771010.pdf}};
      \fill[color=green!20] (0.85, 0.85) rectangle (1.55, 1.55);
  \end{tikzpicture}%

%% file: figs/tasks/noisy_obs_obvector.tikz
\begin{tikzpicture}[font=\small]
    \draw[step=1.0cm,color=gray] (0, 0) grid (5.0, 1.0);
    \node (A1) at (0.5, 0.5) {$0.00$};
    \node (A2) at (1.5, 0.5) {$0.00$};
    \node (A3) at (2.5, 0.5) {$-1.27$};
    \node (A4) at (3.5, 0.5) {$...$};
    \node (A5) at (4.5, 0.5) {$0.10$};

    \draw [->] (2.5, -0.2) -- node[right, font=\footnotesize]{move down} (2.5, -0.8);

    \draw[step=1.0cm,color=gray] (0, -2) grid (5.0, -1);
    \node (A1) at (0.5, -1.5) {$0.00$};
    \node (A2) at (1.5, -1.5) {$1.00$};
    \node (A3) at (2.5, -1.5) {$0.32$};
    \node (A4) at (3.5, -1.5) {$...$};
    \node (A5) at (4.5, -1.5) {$1.45$};

    % \node (1) {[0., 0., -1.27, 0.91, ... , 0.10]};
    % \node (2) [below=of 1] {[0.00,  1.00, -0.01, -1.32, ... , 1.45]};
    % \draw[->] (1) -- node[right]{move down} (2);
\end{tikzpicture}

%% file: figs/tasks/branching.tikz
\begin{tikzpicture}[node distance = 0.2cm and 0.8cm, thick, minimum size=0.5cm]% 
    \node (0) [circ, inner sep=0cm] {\includegraphics[width=0.35cm]{figs/icons/noun_robot_head_771010.pdf}};
    \node (5) [circ, right=of 0] {};
    \node (1) [circ, above=of 5] {};
    \node (2) [circ, above=of 1] {};
    \node (3) [circ, below=of 5] {};
    \node (4) [circ, below=of 3] {};

    % \draw[->] (1) edge[out=0,in=60,looseness=8] (1);
    % \draw[->] (2) edge[out=0,in=60,looseness=8] (2);
    % \draw[->] (3) edge[out=0,in=300,looseness=8] (3);
    % \draw[->] (4) edge[out=0,in=300,looseness=8] (4);

    \draw[->] (0) -- (1);
    \draw[->] (0) -- (2);
    \draw[->] (0) -- (3);
    \draw[->] (0) -- (4);
    \draw[->] (0) -- (5);

    \node (ell) [right=of 5] {...};
    \node (15) [circ, right=of ell] {};
    \node (11) [circ, above=of ell] {};
    \node (12) [circ, above=of 11] {};
    \node (13) [circ, below=of ell] {};
    \node (14) [circ, below=of 13] {};

    \draw[->] (5) -- (11);
    \draw[->] (5) -- (12);
    \draw[->] (5) -- (13);
    \draw[->] (5) -- (14);
    \draw[->] (5) -- (ell);

    \draw[->] (ell) -- (15);
    
    \node (20) [circ, inner sep=0cm, right=of 15, fill=green!30] {};
    \node (16) [circ, above=of 20] {};
    \node (17) [circ, above=of 16] {};
    \node (18) [circ, below=of 20] {};
    \node (19) [circ, below=of 18] {};

    \draw[->] (15) -- (16);
    \draw[->] (15) -- (17);
    \draw[->] (15) -- (18);
    \draw[->] (15) -- (19);
    \draw[->] (15) -- (20);
\end{tikzpicture}%

%% file: figs/tasks/parabola.tikz
\begin{tikzpicture}
  \draw (-0.5, 0) -- (1.8, 0);
  \draw (0, -1.0) -- (0, 1.0);

  \draw[scale=1,domain=-0.1:1.5,smooth,variable=\x,blue!60, thick] plot ({\x},{0.3 + \x - \x*\x});

  \draw[red!70, thick] (0.0, 0.3) -- (0.1, 0.50) -- (0.2, 0.40) -- (0.3, 0.55) -- (0.4, .57) -- (0.5, .61) -- (0.6, .68) -- (0.7, .60) -- (0.8, .55) -- (0.9, .49) -- (1.0, .22);

  \draw[black!70, fill=black!70] (0.0, 0.30) circle (0.02);
  \draw[black!70, fill=black!70] (0.1, 0.50) circle (0.02);
  \draw[black!70, fill=black!70] (0.2, 0.40) circle (0.02);
  \draw[black!70, fill=black!70] (0.3, 0.55) circle (0.02);
  \draw[black!70, fill=black!70] (0.4, 0.57) circle (0.02);
  \draw[black!70, fill=black!70] (0.5, 0.61) circle (0.02);
  \draw[black!70, fill=black!70] (0.6, 0.68) circle (0.02);
  \draw[black!70, fill=black!70] (0.7, 0.60) circle (0.02);
  \draw[black!70, fill=black!70] (0.8, 0.55) circle (0.02);
  \draw[black!70, fill=black!70] (0.9, 0.49) circle (0.02);
  \draw[black!70, fill=black!70] (1.0, 0.22) circle (0.02);
\end{tikzpicture}

%% file: figs/tasks/largest_sum.tikz
\begin{tikzpicture}
    \fill[blue!60!white] (0,0) rectangle (1.0, 1.0);
    \fill[blue!60!white] (1,0) rectangle (2.0, 1.0);
    \fill[orange!60!white] (3,0) rectangle (4.0, 1.0);
    \fill[orange!60!white] (4,0) rectangle (5.0, 1.0);

    \node (A1) at (0.5, 0.5) {$x_1$};
    \node (A2) at (1.5, 0.5) {$x_2$};
    \node (A3) at (2.5, 0.5) {...};
    \node (A4) at (3.5, 0.5) {$x_{2L-1}$};
    \node (A5) at (4.5, 0.5) {$x_{2L}$};

    \fill[blue!60!white] (0.8,-2.2) rectangle (2.2, -0.8);
    \fill[orange!60!white] (2.8,-2.2) rectangle (4.2, -0.8);

    \node (B2) at (1.5, -1.5) {$\displaystyle \sum_{j=1}^L s_j$};
    \node (B3) at (2.5, -1.5) {$\leq$};
    \node (B4) at (3.5, -1.5) {$\displaystyle \sum_{j=L+1}^{2L} s_j$};
    \node (B5) at (4.5, -1.5) {?};
\end{tikzpicture}

%% file: figs/tasks/init_state_shift.tikz
\begin{tikzpicture}[node distance = 0.2cm and 0.8cm, thick, minimum size=0.6cm, font=\footnotesize, inner sep=0cm]% 
    % \node (0) [circ, inner sep=0cm] {\includegraphics[width=0.35cm, decodearray={0 0 1 1 1 1}]{figs/icons/noun_robot_head_771010.pdf}};
    % \node (1) [circ, right=of 0, yshift=+0.8cm] {};
    \node (0) [circ, inner sep=0cm] {$s_{0}$};
    \node (1) [circ, right=of 0, yshift=+0.8cm] {$s_{0}^{*}$};
    \node (2) [circ, right=of 0, yshift=-0.8cm] {};

    \draw[->] (0) -- (1);
    \draw[->] (0) -- (2);

    \node (3) [circ, right=of 1, yshift=+0.4cm, fill=green!15] {};
    \node (4) [circ, right=of 1, yshift=-0.4cm, fill=red!15] {};

    \draw[->] (1) -- (3);
    \draw[->] (1) -- (4);

    \node (5) [circ, right=of 2, yshift=+0.4cm, fill=red!15] {};
    \node (6) [circ, right=of 2, yshift=-0.4cm, fill=green!40] {};

    \draw[->] (2) -- (5);
    \draw[->] (2) -- (6);

    \node (rew-3) [right=of 3, xshift=-0.75cm] {$r = +1.0$};
    \node (rew-4) [right=of 4, xshift=-0.75cm] {$r = -1.0$};
    \node (rew-5) [right=of 5, xshift=-0.75cm] {$r = -1.0$};
    \node (rew-6) [right=of 6, xshift=-0.75cm] {$r = +2.0$};

\end{tikzpicture}%

%% file: figs/tasks/procedural_goal.tikz
 \begin{tikzpicture}[node distance = 1.5cm, thick]% 
      \draw[step=0.4cm,color=gray] (-2.2, -2.2) grid (2.2, 2.2);

      \node[color=gray, opacity=1.0] (A) at (1.4, -1.0) {\includegraphics[width=0.32cm]{figs/icons/noun_robot_head_771010.pdf}};
      \fill[color=green!40] (-0.85, 0.85) rectangle (-1.15, 1.15);
  \end{tikzpicture}%

%% file: figs/tasks/sort.tikz
\begin{tikzpicture}
    \fill[blue!80!white] (3,0) rectangle (4.0, 1.0);
    \fill[blue!60!white] (0,0) rectangle (1.0, 1.0);
    \fill[blue!40!white] (2,0) rectangle (3.0, 1.0);
    \fill[blue!20!white] (1,0) rectangle (2.0, 1.0);

    \node (A1) at (3.5, 0.5) {$0.17$};
    \node (A2) at (0.5, 0.5) {$0.31$};
    \node (A3) at (2.5, 0.5) {$0.54$};
    \node (A4) at (1.5, 0.5) {$0.80$};

    \draw [->] (0.5, -0.2) -- (3.5, -0.8);
    \draw [->] (3.5, -0.2) -- (0.5, -0.8);

    \fill[blue!80!white] (0,-2) rectangle (1.0, -1.0);
    \fill[blue!60!white] (3,-2) rectangle (4.0, -1.0);
    \fill[blue!40!white] (2,-2) rectangle (3.0, -1.0);
    \fill[blue!20!white] (1,-2) rectangle (2.0, -1.0);

    \node (B1) at (0.5, -1.5) {$0.17$};
    \node (B2) at (3.5, -1.5) {$0.31$};
    \node (B3) at (2.5, -1.5) {$0.54$};
    \node (B4) at (1.5, -1.5) {$0.80$};

    \draw [->] (1.5, -2.2) -- (3.5, -2.8);
    \draw [->] (3.5, -2.2) -- (1.5, -2.8);

    \fill[blue!80!white] (0,-4) rectangle (1.0, -3.0);
    \fill[blue!60!white] (1,-4) rectangle (2.0, -3.0);
    \fill[blue!40!white] (2,-4) rectangle (3.0, -3.0);
    \fill[blue!20!white] (3,-4) rectangle (4.0, -3.0);

    \node (C1) at (0.5, -3.5) {$0.17$};
    \node (C2) at (1.5, -3.5) {$0.31$};
    \node (C3) at (2.5, -3.5) {$0.54$};
    \node (C4) at (3.5, -3.5) {$0.80$};

    % \node (2) [below=of 1]{[0.154, 0.742, 0.946, 0.118]};
    % \draw (1) [->] -- node [right]{(1, 3)} (2);
\end{tikzpicture}

%% file: experiments.tex
\section{Benchmarking Algorithms}
\label{sec:experiments}

\begin{figure}
    \centering
    \includegraphics[]{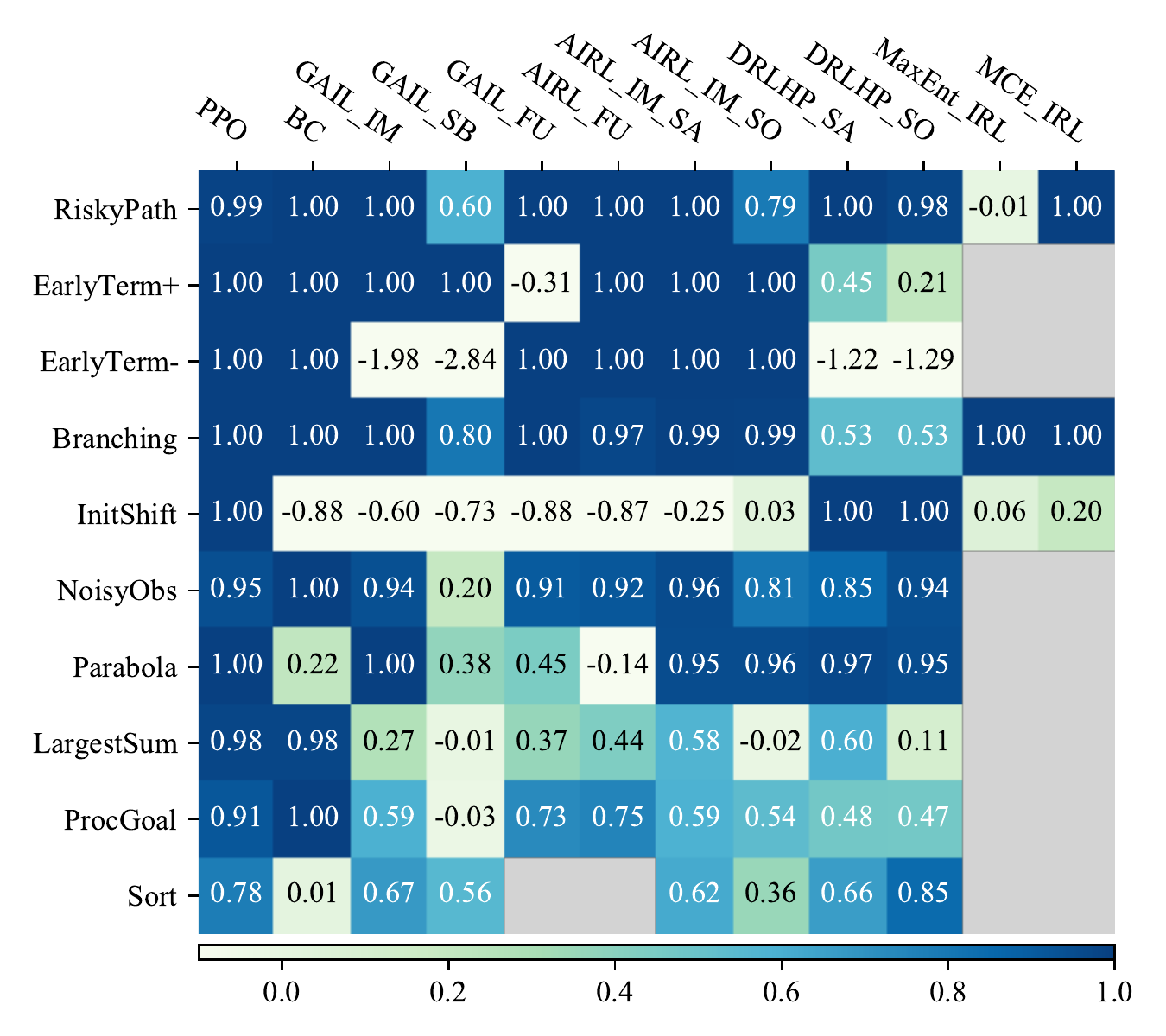}
    \caption{Mean episode return (across 15 seeds) of policy learned by each algorithm ($x$-axis) on each task ($y$-axis).
    Returns are normalized between $0.0$ for a random policy and $1.0$ for an optimal policy.
        Grey cells denote the algorithm being unable to run on that task (e.g. \maxent{} and \mce{} only run on small tabular tasks). See appendix~\ref{sec:appendix:experimental-results} for full results and confidence intervals.}
    \label{fig:experiments:heatmap}
\end{figure}

\subsection{Experimental Setup}

We evaluate a range of commonly used LfH algorithms: Maximum Entropy IRL~(\maxent{}; \citealp{ziebart:2008}),
Maximum Causal Entropy IRL~(\mce{}; \citealp{ziebart:2010:thesis}), Behavioral
Cloning (\bc{}), Generative Adversarial Imitation Learning~(\gail{}; \citealp{ho:2016}), Adversarial
IRL~(\airl{}; \citealp{fu:2018}) and Deep Reinforcement Learning from Human
Preferences (\drlhp{}; \citealp{christiano:2017}). 
We also present an RL baseline using Proximal Policy Optimization~(\ppo{}; \citealp{schulman:2017}).

We test several variants of these algorithms.
We compare multiple implementations of \airl{} (\airlim{} and \airlfu{}) and \gail{} (\gailim{}, \gailfu{} and \gailsb{}).
We also vary whether the reward function in \airlim{} and \drlhp{} is state-only (\airlimso{} and \drlhpso{}) or state-action (\airlimsa{} and \drlhpsa{}).
All other algorithms use state-action rewards.

We train \drlhp{} using preference comparisons from the ground-truth reward, and train all other algorithms using demonstrations from an optimal expert policy.
We compute the expert policy using value iteration in discrete environments, and procedurally specify the expert in other environments.
See appendix~\ref{sec:appendix:experimental-setup} for a complete description of the experimental setup and implementations.

\subsection{Experimental Results}
\label{sec:experiments:results}

For brevity, we highlight a few notable findings, summarizing our results in Figure~\ref{fig:experiments:heatmap}.
See appendix~\ref{sec:appendix:experimental-results} for the full results and a more comprehensive analysis.

\textbf{Implementation Matters}.
Our results show that \gailim{} and \gailsb{} are biased towards prolonging episodes: they achieve \emph{worse than random} return on \earlyn{}, where the optimal action is to end the episode, but \emph{match expert performance} in \earlyp{}.
By contrast, \gailfu{} is biased towards ending the episode, succeeding in \earlyn{} but failing in \earlyp{}.
This termination bias can be a major confounder when evaluating on variable-horizon tasks.

We also observe several other differences between implementations of the same algorithm.
\gailsb{} achieves significantly lower return than \gailim{} and \gailfu{} on \noisyobs{}, \parabola{}, \largest{} and \procgoal{}.
Moreover, \airlim{} attains near-expert return on \parabola{} while \airlfu{} performs worse than random.
These results confirm that implementation matters~\citep{islam2017reproducibility,henderson2018deep,engstrom:2020},
and illustrate how diagnostic tasks can pinpoint how performance varies between implementations.

\textbf{Rewards vs Policy Learning}.
Behavioral cloning (\bc{}), fitting a policy to demonstrations using supervised learning, exhibits bimodal performance.
\bc{} often attains near-optimal returns.
However, in tasks with large continuous state spaces such as \parabola{} and \sort{}, \bc{} performs close to random.
We conjecture this is because \bc{} has more difficulty generalizing to novel states than reward learning algorithms, which have an inductive bias towards goal-directed behavior.

\begin{figure}
    \centering
    \includegraphics[trim=3 0 0 0, clip, width=0.7\textwidth]{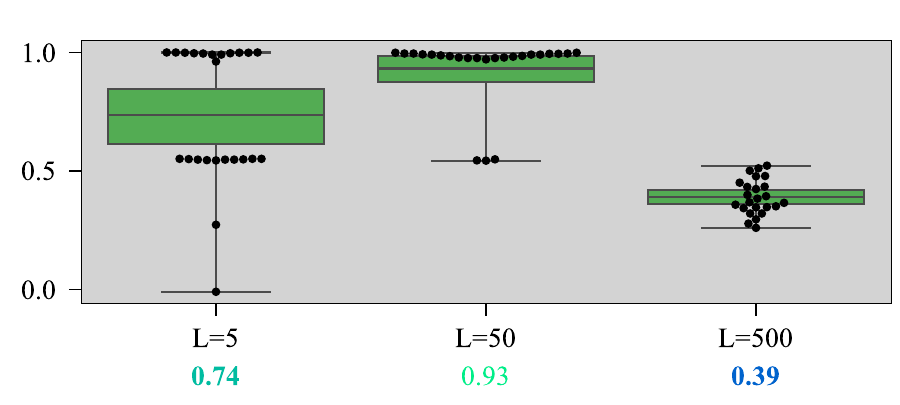}
    \caption{\drlhpsa{} return on \noisyobs{} for varying numbers of noise
      dimensions $L$ (grid size $M=7$). We evaluate across 24 seeds trained for
      3M timesteps. Mean returns are depicted as horizontal lines inside the boxes and reported
      underneath $x$-axis labels. Boxes span the 95\% confidence intervals of the means; whiskers span the range of returns.}
    \label{fig:experiments:drlhpnoise}
\end{figure}

\textbf{Exploration in Preference Comparison}.
\label{drlhp-exploration}
We find \drlhp{}, which learns from preference comparisons, achieves lower returns in \branch{} than algorithms that learn from demonstrations.
This is likely because \branch{} is a hard-exploration task: a specific sequence of actions must be taken in order to achieve a reward.
For \drlhp{} to succeed, it must first discover this sequence, whereas algorithms that learn from demonstrations can simply mimic the expert.

While this problem is particularly acute in \branch{}, exploration is likely to limit the performance of \drlhp{} in other environments.
To investigate this further, we varied the number of noise dimensions $L$ in \noisyobs{} from $5$ to $500$, reporting the performance of \drlhpsa{} in Figure~\ref{fig:experiments:drlhpnoise}.
Increasing $L$ decreases both the maximum and the variance of the return.
This causes a \emph{higher} mean return in $L=50$ than in $L=5$ (high variance) or $L=500$ (low maximum).

We conjecture this behavior is partly due to \drlhp{} comparing trajectories sampled from a policy optimized for its current best-guess reward.
If the policy becomes low-entropy too soon, then \drlhp{} will fail to sufficiently explore.
Adding stochasticity stabilizes the training process, but makes it harder to recover the true reward.

%% file: case_study.tex
\section{Case Study: Improving Implementations}
\label{sec:case-study}
In the previous section, we showed how DERAIL can be used to compare existing implementations of reward and imitation learning algorithms.
However, benchmarks are also often used during the development of new algorithms and implementations.
We believe diagnostic task suites are particularly well-suited to rapid prototyping.
The tasks are lightweight so tests can be conducted quickly.
Yet they are sufficiently varied to catch a wide range of bugs, and give a glimpse of effects in more complex environments.
To illustrate this workflow, we present a case study refining an implementation of Deep Reinforcement Learning from Human Preferences~(\citealp[][\drlhp{}]{christiano:2017}).

As discussed in section~\ref{sec:experiments:results}, the implementation we experimented with, \drlhp{}, has high-variance across random seeds.
We conjecture this problem occurs because the preference queries are insufficiently diverse.
The queries are sampled from rollouts of a policy, and so their diversity depends on the stochasticity of the environment and policy.
Indeed, we see in Figure~\ref{fig:experiments:drlhpnoise} that \drlhp{} is more stable when environment stochasticity increases.

The fundamental issue is that \drlhp{}'s query distribution depends on the policy, which is being trained to maximize \drlhp{}'s \emph{predicted} reward.
This entanglement makes the procedure liable to get stuck in local minima.
Suppose that, mid-way through training, the policy chances upon some previously unseen, high-reward state.
The predicted reward at this unseen state will be random -- and so likely worse than a previously seen, medium-reward state.
The policy will thus be trained to \emph{avoid} this high-reward state -- starving \drlhp{} of the queries that would allow it to learn in this region.

In an attempt to address this issue, we experiment with a few simple modifications to \drlhp{}:
\label{dlrhp-mod-ideas}
\begin{itemize}
  \item \drlhpa{}. Reduce the learning rate for the policy. The policy is initially high-entropy; over time, it learns to only take actions with high predicted reward. By slowing down policy learning, we maintain a higher-entropy query distribution.
  \item \drlhpb{}. When sampling trajectories for preference comparison, use an $\epsilon$-greedy version of the current policy (with $\epsilon = 0.1$). This directly increases the entropy of the query distribution.
  \item \drlhpc{}. Add an exploration bonus using random network distillation~\citep{burda2018exploration}. Distillation steps are performed only on trajectories submitted for preference comparison. This has the effect of giving a bonus to state-action pairs that are uncommon in preference queries (even if they occurred frequently during policy training).
\end{itemize}

\begin{figure}
    \centering
    \includegraphics[]{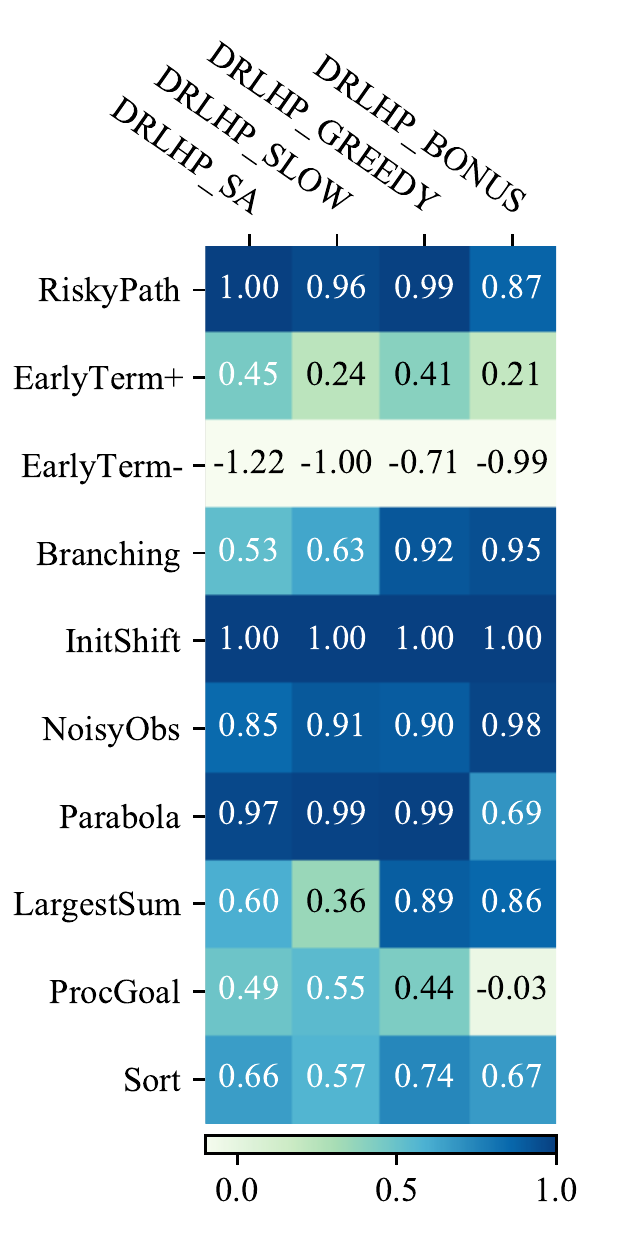}
    \caption{Return of \drlhpsa{} and three variants: \drlhpa{}, slower policy training; \drlhpb{}, $\epsilon$-greedy exploration; \drlhpc{}, exploration bonus (see section~\ref{dlrhp-mod-ideas}). Mean episode return (across 15 seeds) of policy learned by each algorithm ($x$-axis) on each task ($y$-axis).
    Returns are normalized between $0.0$ for a random policy and $1.0$ for an optimal policy.}
    \label{fig:experiments:drlhp-mods}
\end{figure}
We report the return achieved with these modifications in Figure~\ref{fig:experiments:drlhp-mods}.
\drlhpb{} produces the most stable results: the returns are all comparable to or substantially higher than the original \drlhpsa{}.
However, all modifications increase returns on hard-exploration task \branch{}, although for \drlhpa{} the improvement is modest.
\drlhpb{} and \drlhpc{} also enjoy significant improvements on high-dimensional classification task \largest{}, which likely benefits from more balanced labels.
\drlhpc{} performs poorly on \parabola{} and \procgoal{}: we conjecture that the large state space caused \drlhpc{} to explore excessively.

This case study shows how DERAIL can help rapidly test new prototypes, quickly confirming or discrediting a hypothesis of a how a change will affect a given algorithm.
Moreover, we can gain a fine-grained understanding of performance along different axes.
For example, we could conclude that \drlhpc{} does increase exploration (higher return on \branch{}) but may over-explore (lower return on \procgoal{}).
It would be difficult to disentangle these distinct effects in more complex environments.

%% file: discussion.tex
\section{Discussion}

We have developed, to the best of our knowledge, the first suite of diagnostic environments for reward and imitation learning algorithms.
We find that by isolating particular algorithmic capabilities, diagnostic tasks can provide a more nuanced picture of individual algorithms' strengths and weaknesses than testing on more complex benchmarks.
Our results confirm that reward and imitation learning algorithm performance is highly sensitive to implementation details.
Furthermore, we have demonstrated the fragility of behavioral cloning, and obtained qualitative insights into the performance of preference-based reward learning.
Finally, we have illustrated in a case study how DERAIL can support rapid prototyping of algorithmic refinements.

In designing the task suite, we have leveraged our personal experience as well as past work documenting design flaws and implementation bugs~\citep{ziebart:2010:thesis,kostrikov2018discriminator}.
We expect to refine and extend the suite in response to user feedback, and we encourage other researchers to develop complementary tasks.
Our environments are open-source and available at \projectsource{}.

%% file: task_specs.tex
\renewcommand{\labelitemii}{$\circ$}
\newcommand{\tikzsidefigure}[3][1.0]{
    \begin{center}
        \captionsetup{type=figure}
        \scalebox{#1}{\input{figs/tasks/#2.tikz}}
        \captionof{figure}{#3}
    \end{center}
}
\newcommand{\taskdescwidth}{0.53\textwidth}
\newcommand{\taskfigwidth}{0.37\textwidth}
\newcommand{\taskcolmargin}{0.1\textwidth}
\newcommand{\taskcolumns}[2]{%
\begin{minipage}{\taskdescwidth}
#1
\end{minipage}%
\hspace{\taskcolmargin}
\begin{minipage}{\taskfigwidth}
#2
\end{minipage}
}

\section{Full specification of tasks}
\label{sec:appendix:tasks-specification}

\subsection{\risky{}}

\taskcolumns{
\begin{itemize}
  \item State space: $\statespace{} = \{0, 1, 2, 3\}$
  \item Action space: $\actionspace{} = \{0, 1\}$
  \item Horizon: $5$
  \item Dynamics:
  \begin{itemize}
    \item $0 \xrightarrow[]{0} 1$, $0 \xrightarrow[]{1} \begin{cases}1, & \text{50\% probability;} \\ 2, & \text{50\% probability.}\end{cases}$
    \item $1 \xrightarrow[]{0} 2$, $1 \xrightarrow[]{1} 1$
    \item $2 \xrightarrow[]{a} 2$ for all $a \in \actionspace{}$
    \item $3 \xrightarrow[]{a} 3$ for all $a \in \actionspace{}$
  \end{itemize}
  \item Rewards: $R(0) = R(1) = 0, R(2) = 1, R(3) = -100$
\end{itemize}
}{
    \begin{center}
        \captionsetup{type=figure}
        \scalebox{1.0}{\input{figs/tasks/risky_path.tikz}}
        \captionof{figure}{\risky{}}
    \end{center}
}

\subsection{\earlypn{}}

\taskcolumns{
\begin{itemize}
  \item State space: $\statespace{} = \{0, 1, 2\}$
  \item Action space: $\actionspace{} = \{0, 1\}$
  \item Horizon: $10$
  \item Dynamics:
  \begin{itemize}
    \item $0 \xrightarrow[]{a} 1$ for all $a \in \actionspace{}$
    \item $1 \xrightarrow[]{0} 0$, $1 \xrightarrow[]{1} 2$
  \end{itemize}
  \item Rewards: $R(s) = 1.0$ in \earlyp{} \\and $R(s) = -1.0$ in \earlyn{}
\end{itemize}
}{
  
    \begin{center}
        \captionsetup{type=figure}
        \scalebox{1.0}{\input{figs/tasks/early_termination_pos.tikz}}
        \captionof{figure}{\earlyp{}}
    \end{center}

  \vspace{10pt}

    \begin{center}
        \captionsetup{type=figure}
        \scalebox{1.0}{\input{figs/tasks/early_termination_neg.tikz}}
        \captionof{figure}{\earlyn{}}
    \end{center}

}

\subsection{\noisyobs{}}

\taskcolumns{
\begin{itemize}
  \item Configurable parameters: size $M=5$, noise length $L=20$
  \item State space: grid $\statespace{} = \mathbb{Z}_M \times \mathbb{Z}_M$, where $\mathbb{Z}_M = \{0, 1, \cdots, M-1\}$
  \item Observations: $(x,y,z_1,\cdots,z_L)$, where:
    \begin{itemize}
      \item $(x, y) \in \statespace{}$ are the state coordinates
      \item $z_i$ are i.i.d.\ samples from the Gaussian $\mathcal{N}(0, 1)$
    \end{itemize}
  \item Action space: $\actionspace{} = \{U,D,L,R,S\}$ for moving Up, Down, Left, Right or Stay (no-op).
  \item Horizon: $3M$
  \item Dynamics: Deterministic gridworld; attempting to move beyond boundary of world is a no-op.
  \item Rewards: $R(s) = \mathbbm{1}\left[s = \left(\floor{\frac{M}{2}}, \floor{\frac{M}{2}}\right)\right]$
  \item Initial state: $(0, 0)$
\end{itemize}
}{
\begin{center}
    \captionsetup{type=figure}
    \input{figs/tasks/noisy_obs_states.tikz}

    \vspace{10pt}

    \scalebox{0.8}{
        \input{figs/tasks/noisy_obs_obvector.tikz}
    }
    \captionof{figure}{\noisyobs{}}
\end{center}
}

\subsection{\branch{}}

\taskcolumns{
\begin{itemize}
  \item Configurable parameters: path length $L=10$, branching factor $B=2$
  \item State space: $\statespace{} = \{0, 1, \cdots, LB\}$
  \item Action space: $\actionspace{} = \{0, \cdots, B-1\}$
  \item Horizon: $L$
  \item Dynamics: $s \xrightarrow[]{a} s +  (a + 1) \cdot \mathbbm{1}[s \equiv 0 \mod B]$
  \item Rewards: $R(s) = \mathbbm{1}\left[s = LB \right]$
  \item Initial state: $0$
\end{itemize}
}{
    \begin{center}
        \captionsetup{type=figure}
        \scalebox{1.0}{\input{figs/tasks/branching.tikz}}
        \captionof{figure}{\branch{}}
    \end{center}
}

\subsection{\parabola{}}

\taskcolumns{
\begin{itemize}
  \item Configurable parameters: x-step $dx=0.05$, horizon $h=20$
  \item State space: $\statespace{} = \mathbbm{R}^2 \times [-1, 1]^3$
  \item Actions: $a \in (-\infty, +\infty)$
  \item Horizon: $h$
  \item Dynamics: \[\left((x, y), (c_2, c_1, c_0)\right) \xrightarrow[]{a} \left((x + dx, y + a), (c_2, c_1, c_0)\right)\]
  \item Rewards: $R(s) = (c_2 x^2 + c_1 x + c_0 - y)^2$
  \item Initial state: $(0, c_0, c_0, c_1, c_2)$, where $c_i \sim \operatorname{Unif}([-1, 1])$ 
\end{itemize}
}{
    \begin{center}
        \captionsetup{type=figure}
        \scalebox{1.8}{\input{figs/tasks/parabola.tikz}}
        \captionof{figure}{\parabola{}}
    \end{center}
}

\subsection{\largest{}}

\taskcolumns{
\begin{itemize}
  \item Configurable parameters: half-length $L=25$
  \item State space: $\statespace{} = [0, 1]^{2L}$
  \item Action space: $\actionspace{} = \{0, 1\}$
  \item Horizon: $1$
  \item Rewards: $R(s, a) = \mathbbm{1}\left[a = \mathbbm{1}\left[\sum_{j=1}^L s_j \leq \sum_{j=L+1}^{2L} s_j\right]\right]$
  \item Initial state: $s_0 \sim \operatorname{Unif}\left([0, 1]^{2L}\right)$
\end{itemize}
}{
    \begin{center}
        \captionsetup{type=figure}
        \scalebox{0.9}{\input{figs/tasks/largest_sum.tikz}}
        \captionof{figure}{\largest{}}
    \end{center}
}

\subsection{\initshift{}}

\taskcolumns{
\begin{itemize}
  \item State space: $\statespace{} = \{0, 1, .., 6\}$
  \item Action space: $\actionspace{} = \{0, 1\}$
  \item Horizon: $2$
  \item Dynamics:
    $s \xrightarrow[]{a} 2 s + 1 + a$
  \item Rewards:
    \[
      R(s) = \begin{dcases*}
        +1 & if $s = 3$\\
        -1 & if $s \in \{4, 5\}$\\
        +2 & if $s = 6$\\
        0  & otherwise \end{dcases*}
    \]
  \item Initial state:\\
    \begin{itemize}
    \item Expert: $0$
    \item Learner: $1$
    \end{itemize}

\end{itemize}
}{
    \begin{center}
        \captionsetup{type=figure}
        \scalebox{1.0}{\input{figs/tasks/init_state_shift.tikz}}
        \captionof{figure}{\initshift{}}
    \end{center}
}

\subsection{\procgoal{}}

\taskcolumns{
\begin{itemize}
  \item Configurable prameters: initial state bound $B=100$, goal distance $D=10$
  \item State space: $\statespace{} = \mathbb{Z}^2 \times \mathbb{Z}^2$, where $(p,g) \in \statespace$ consists of agent position $p$ and goal position $g$
  \item Action space: $\actionspace{} = \{U,D,L,R,S\}$ for moving Up, Down, Left, Right or Stay (no-op).
  \item Horizon: $3D$
  \item Dynamics: Deterministic gridworld on $p$; $g$ is fixed at start of episode.
  \item Rewards: $R((p,g)) = - \lVert p - g \rVert_1$
  \item Initial state:
    \begin{itemize}
    \item Position $p$ uniform over $\{p \in \mathbb{Z}^2 :  \lVert p \rVert_1 \leq B\}$
    \item Goal $g$ uniform over $\{g \in \mathbb{Z}^2  : \lVert p - g \rVert_1 = D\}$
    \end{itemize}
\end{itemize}
}{
    \begin{center}
        \captionsetup{type=figure}
        \scalebox{1.0}{\input{figs/tasks/random_goal.tikz}}
        \captionof{figure}{\procgoal{}}
    \end{center}
}

\subsection{\sort{}}

\taskcolumns{
\begin{itemize}
  \item Configurable parameters: length $L=4$
  \item State space: $\statespace{} = [0, 1]^L$
  \item Action space: $\actionspace{} = \mathbb{Z}_L \times \mathbb{Z}_L$ where $\mathbb{Z}_L = \{0, .., L-1\}$
  \item Horizon: $2L$
  \item Dynamics: $a = (i, j)$ swaps elements $i$ and $j$
  \item Rewards: $R(s, s') = \mathbbm{1}\left[c(s') = n\right] + c(s') - c(s)$, where $c(s)$ is the number of elements in the correct sorted position.
  \item Initial state: $s \sim \operatorname{Unif}([0, 1]^{L})$
\end{itemize}
}{
    \begin{center}
        \captionsetup{type=figure}
        \scalebox{0.8}{\input{figs/tasks/sort.tikz}}
        \captionof{figure}{\sort{}}
    \end{center}
}

%% file: figs/tasks/random_goal.tikz
 \begin{tikzpicture}[node distance = 1.5cm, thick]% 
      \draw[step=0.4cm,color=gray] (-2.2, -2.2) grid (2.2, 2.2);

      \node[color=gray, opacity=1.0] (A) at (1.4, -1.0) {\includegraphics[width=0.32cm]{figs/icons/noun_robot_head_771010.pdf}};
      \fill[color=green!40] (-0.85, 0.85) rectangle (-1.15, 1.15);
  \end{tikzpicture}%

%% file: experiments_full.tex
\section{Experimental setup}
\label{sec:appendix:experimental-setup}

\subsection{Algorithms}

The exact code for running the experiments and generating the plots can be found
at \algorithmsource{}.

Imitation learning and IRL algorithms are trained using rollouts from an optimal policy. 
The number of expert timesteps provided is the same as the number of timesteps each algorithm runs for.
For \drlhp{}, trajectories are compared using the ground-truth reward. 
The trajectory queries are generated from the policy being learned jointly with the reward.

We used open source implementations of these algorithms, as listed in Table~\ref{table:sourcecodes}. We did not perform hyperparameter tuning, and relied on default values for most hyperparameters. 

\newcommand{\imitationsource}{\href{https://github.com/HumanCompatibleAI/imitation}{HumanCompatibleAI/imitation}}
\newcommand{\stablesource}{\href{https://github.com/hill-a/stable-baselines}{hill-a/stable-baselines}}
\newcommand{\evalrewssource}{\href{https://github.com/HumanCompatibleAI/evaluating-rewards}{HumanCompatibleAI/evaluating-rewards}}
\newcommand{\fusource}{\href{https://github.com/justinjfu/inverse_rl}{justinjfu/inverse\_rl}}

\begin{table}
  \centering
\begin{tabular}{|l|l|}
  \hline
  Source Code & Algorithms \\
  \hline
  \hline
  \citet{stable-baselines:2018} & \ppo{}, \bc{}, \gailsb{}  \\
  \hline
  \citet{fu-inverse-rl:2018} & \airlfu{}, \gailfu{}  \\
  \hline
  \citet{imitation:2020} & \airlimsa{}, \airlimso{}, \gailim{}, \mce{}, \maxent{} \\
  \hline
  \citet{evaluating-rewards:2020} & \drlhpsa{}, \drlhpso{}  \\
  \hline
\end{tabular}
\caption{Sources of algorithm implementations (some of which were slightly adapted).
}
\label{table:sourcecodes}
\end{table}

\subsection{Evaluation}

We run each algorithm with 15 different seeds and 500,000 timesteps.
To evaluate a policy, we compute the exact expected episode return in discrete state environments. In other environments, we compute the average return over 1000 episodes.
The score in a task is the mean return of the learned policy, normalized such
that a policy taking random actions gets a score of 0.0 and the expert gets a
score of 1.0. For \earlyn{}, poor policies can reach values smaller than -3.0;
to keep scores in a similar range to other tasks, we truncate negative values at -1.0.

\section{Complete experimental results}
\label{sec:appendix:experimental-results}

We provide results and analysis grouped around individually tasks~(section~\ref{sec:appendix:task-results}) and algorithms~(section~\ref{sec:appendix:algo-results}).
The results are presented using boxplot graphs, such as Figure~\ref{fig:appendix:risky-path}.
The y-axis represents the return of the learned policy, while the $x$-axis contains different algorithms or tasks.
Each point corresponds to a different seed.
The means across seeds are represented as horizontal lines inside the boxes, with the boxes
spanning bootstrapped $95\%$ confidence intervals of the means; the whiskers show the full range of returns.
Each box is assigned a different color to aid in visually distinguishing the tasks; they do not have any semantic meaning.

\newcommand{\taskplot}[1]{\includegraphics[trim=5 0 0 0, clip, width=1.03\textwidth]{figs/results/tasks/#1}}
\newcommand{\algoplot}[1]{\includegraphics[trim=5 0 0 0, clip, width=1.03\textwidth]{figs/results/algos/#1}}

\subsection{Tasks}
\label{sec:appendix:task-results}

\begin{figure}[h!]
    \centering
    \taskplot{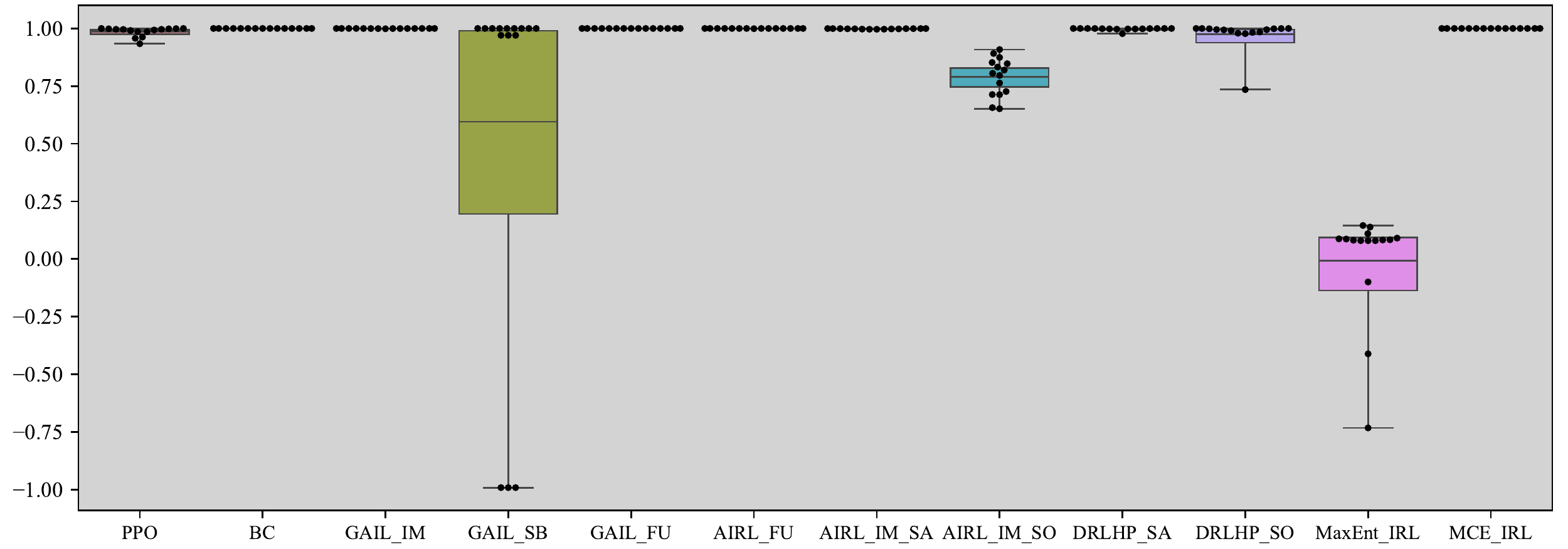}
    \caption{\risky{}: \maxent{} performs poorly as expected, while \mce{} performs well. Other algorithms evaluated look at state-action pairs individually, instead of looking at trajectories, avoiding the problem of risky behavior.}
    \label{fig:appendix:risky-path}
\end{figure}

\begin{figure}[h!]
    \centering
    \begin{subfigure}{\textwidth}
      \taskplot{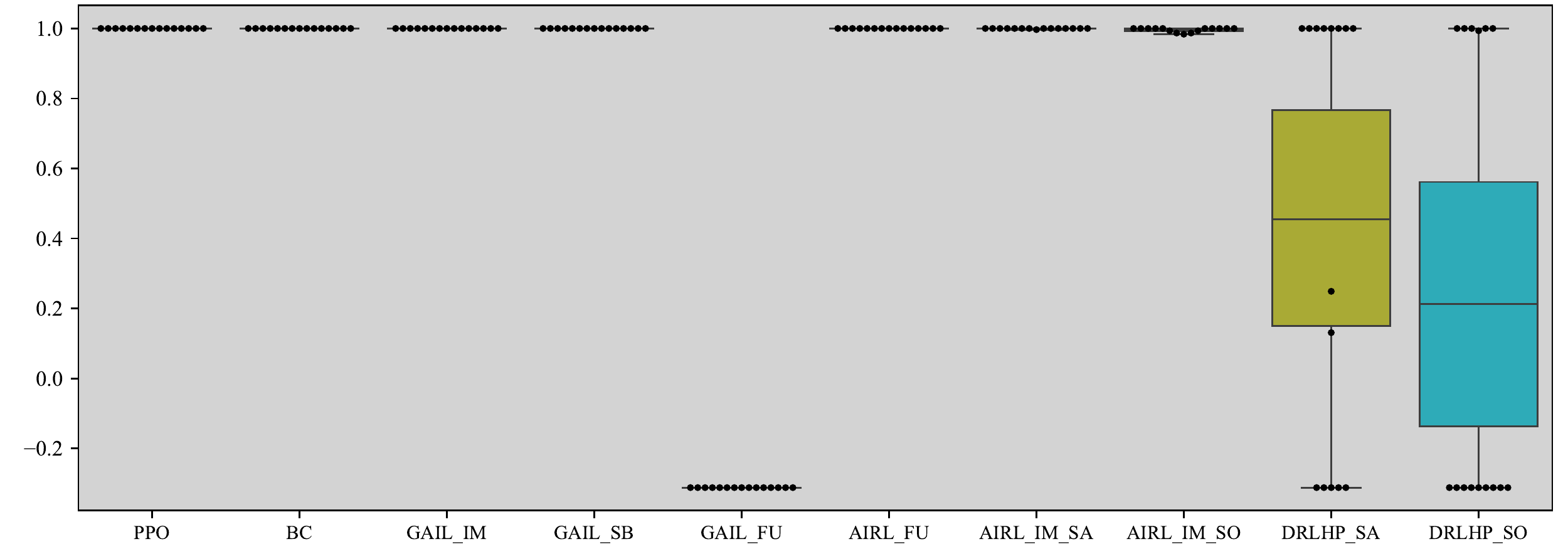}
      \caption{\earlyp}
    \end{subfigure}

    \begin{subfigure}{\textwidth}
      \taskplot{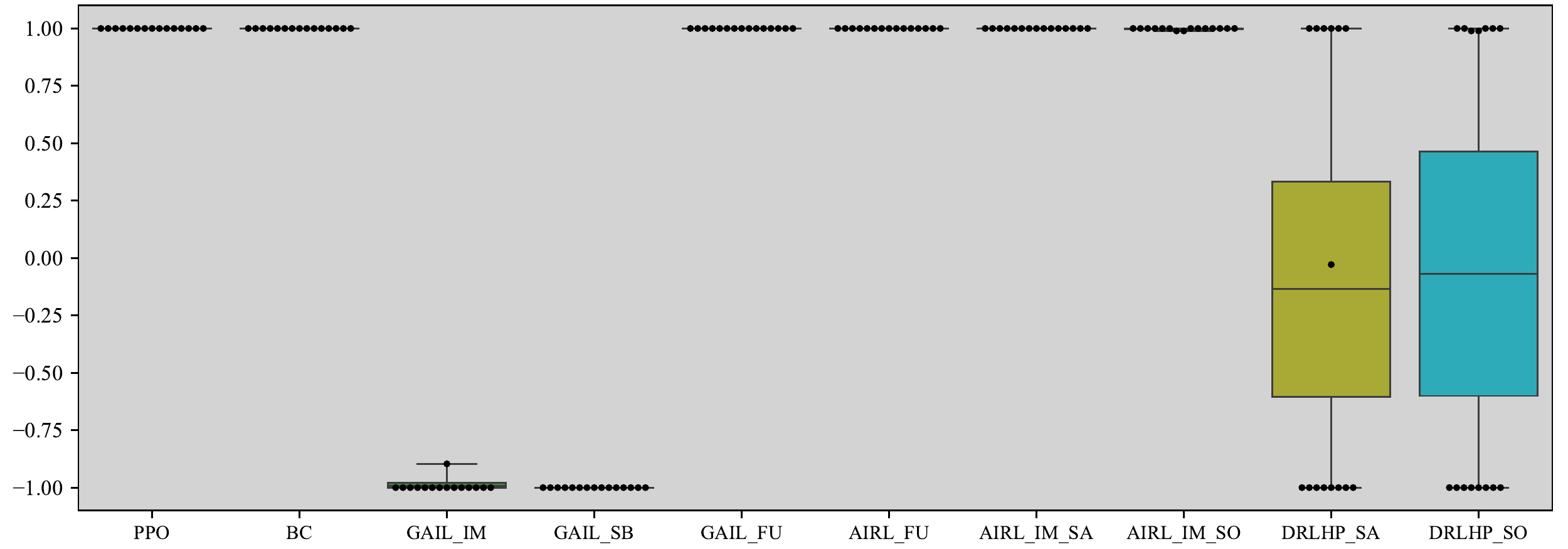}
      \caption{\earlyn}
    \end{subfigure}
    \caption{\earlypn{}: \gailfu{} performs worse than random in \earlyp{} and
      at expert level at \earlyn{}, and we see the opposite behavior with
      \gailim{} and \gailsb{}. This indicates that \gailfu{} has a negatively
      biased reward that favors episode termination, while \gailim{} and
      \gailsb{} have a positively biased reward that favors survival. \drlhp{}
      also has extremely high variance, achieving both expert and random
      performance across different runs. This is because the implementation
      tested looks at segments of trajectories of the same length, without
      accounting for the fact that some segments will cause early termination.
      Instead, every segment is assigned the same reward, and the agent keeps their randomly initialized reward throughout the training process (which might by chance induce expert performance in such a simple environment).
}
\end{figure}

\begin{figure}[h!]
    \centering
  \taskplot{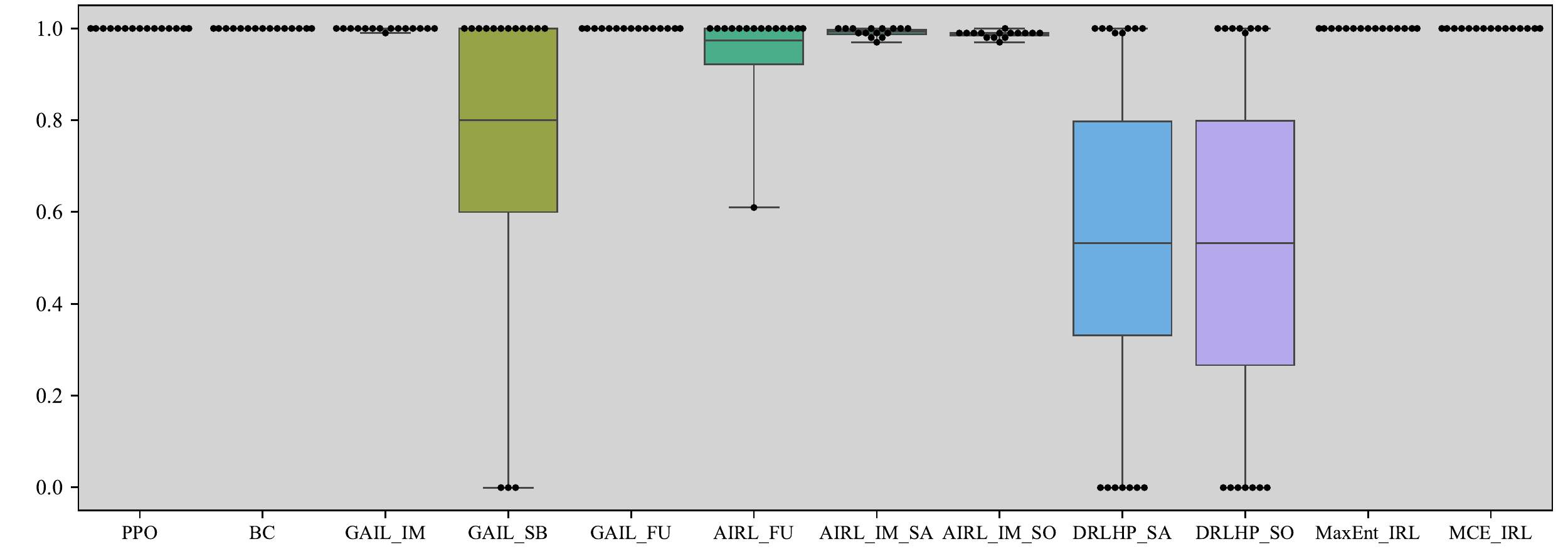}
    \caption{\branch{}: algorithms that learn from expert demonstrations tend
      to perform well, since they require limited exploration. On the other hand,
      \drlhp{} can struggle to perform enough exploration to consistently find
      the goal and learn the correct reward. Note that \drlhp{} needs to find the goal
      multiple times in order to update the reward significantly.
}
\end{figure}

\begin{figure}[h!]
    \centering
  \taskplot{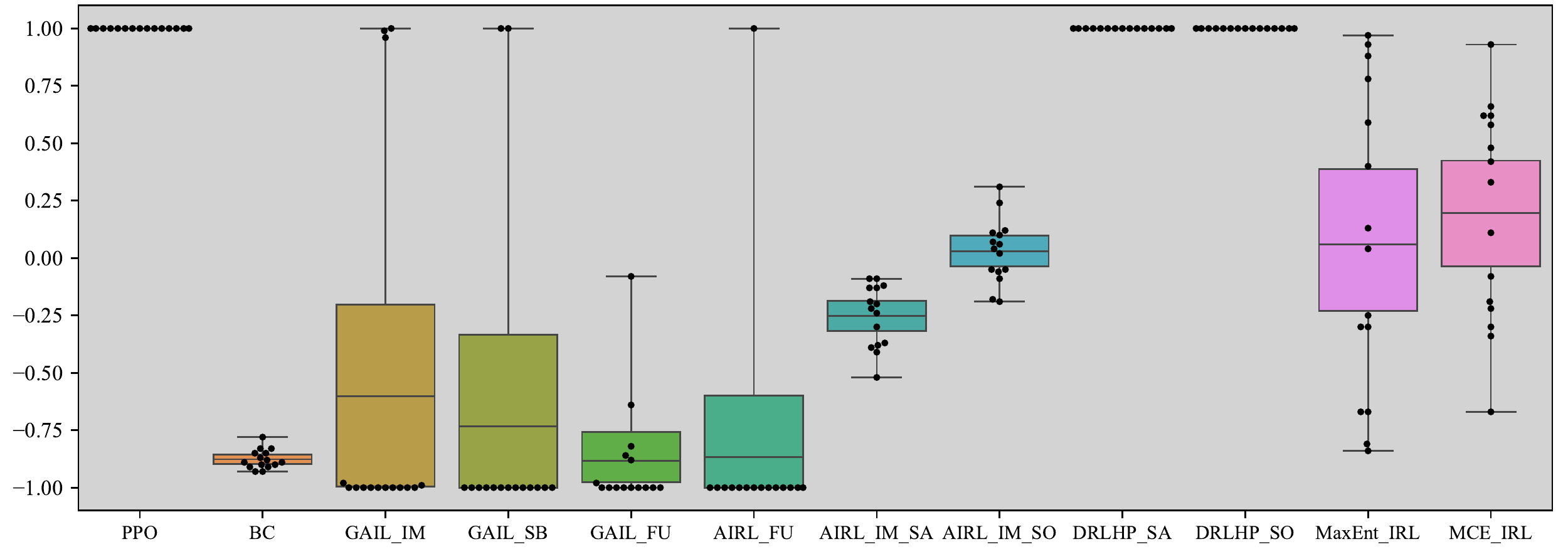}
    \caption{\initstate{}: unlike \branch{}, in \initstate{} algorithms based on expert demonstrations fail, since the expert trajectories do not include the new initial state. By contrast, the task is trivial for \drlhp{}, which can compare trajectories generated at train time. Moreover, algorithms that learn state-action rewards from demonstrations perform \emph{worse} than random. This is because the expert trajectories only contain action 1, and thus rewards tend to assign a positive weight to action $1$. However, the optimal action under the learners initial state distribution is to take action $0$. \airlimso{}, \maxent{} and \mce{} learn state-only reward functions, and perform closer to the random policy.}
\end{figure}

\begin{figure}[h!]
    \centering
    \taskplot{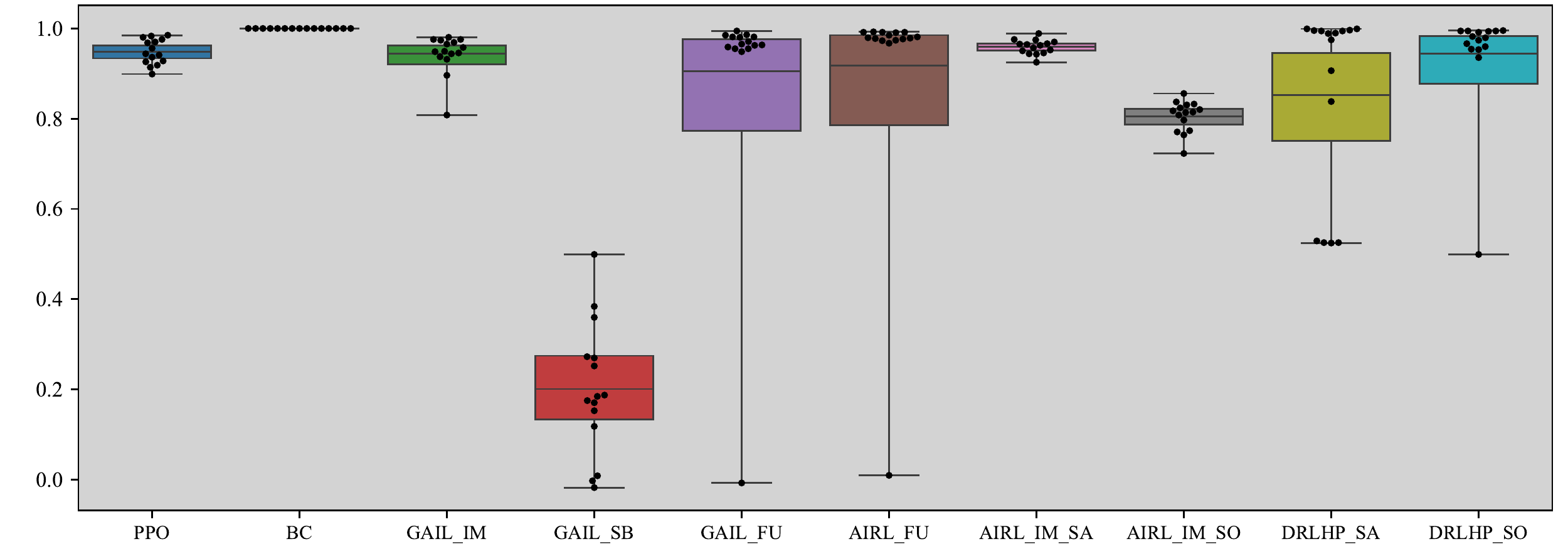}
    \caption{\noisyobs{}: we see that \bc{} achieves near-optimal performance, demonstrating that supervised learning can be more robust and sample-efficient in the presence of noise than other LfH algorithms. We also see that \gailsb{} performs poorly relative to \gailim{} and \gailfu{}, which underscores the importance of the subtle differences between these implementations.}
\end{figure}

\begin{figure}[h!]
    \centering
  \taskplot{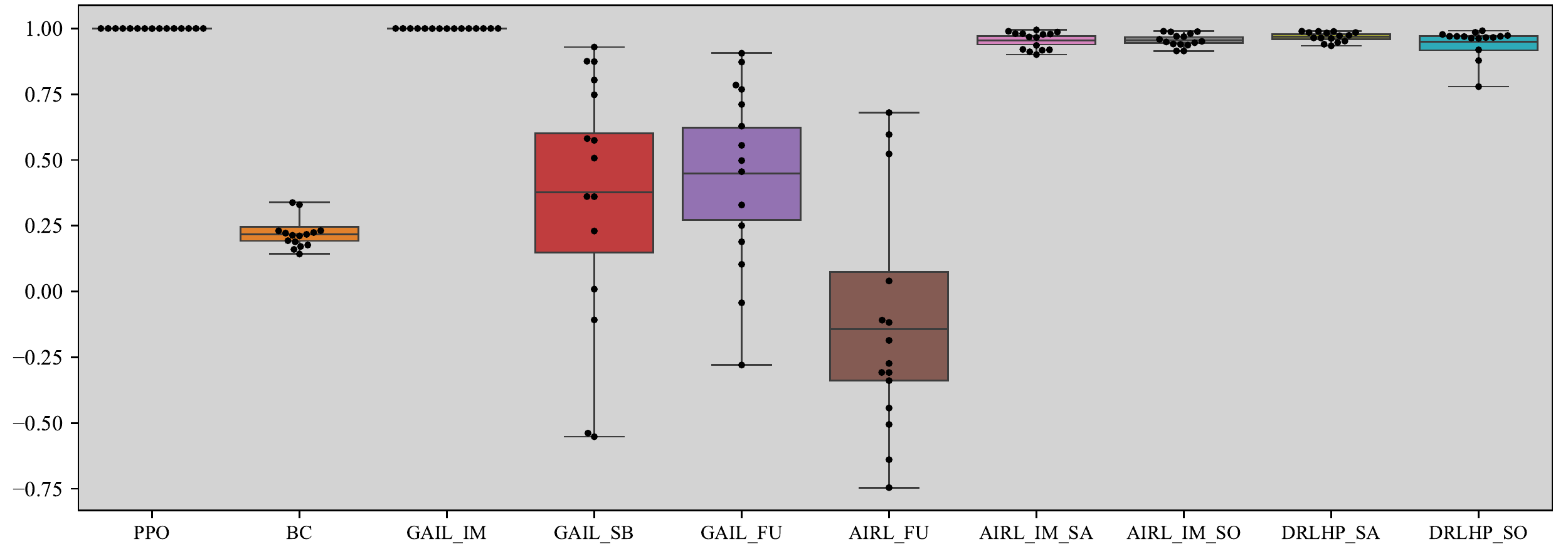}
    \caption{\parabola{}: most algorithms perform well, except for \bc{}, \gailsb{} and \airlfu{}.}
\end{figure}

\begin{figure}[h!]
    \centering
    \taskplot{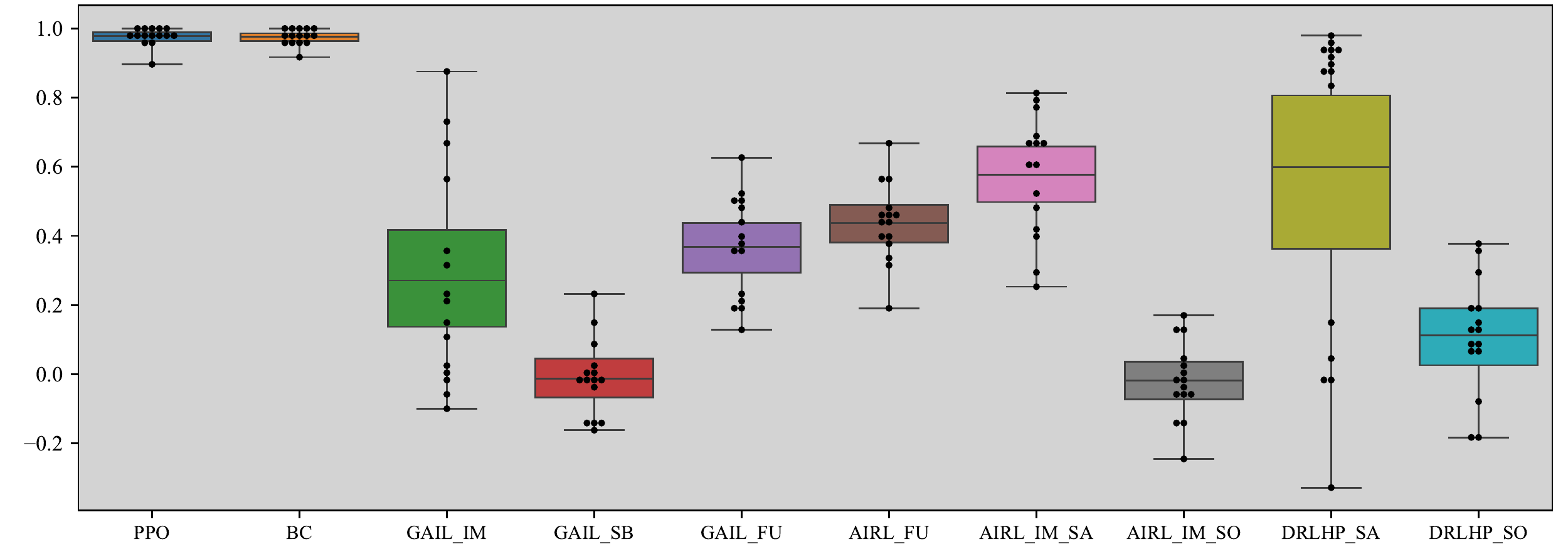}
    \caption{\largest{}: Most algorithms fail to achieve expert performance, while \bc{} does match expert performance, suggesting scaling algorithms like \gail{} and \airl{} to high-dimensional tasks may be a fruitful direction for future work. Methods using state-only reward functions, \airlso{} and \drlhpso{}, perform poorly since the reward for this task depends on the actions taken.}
\end{figure}

\begin{figure}[h!]
    \centering
    \taskplot{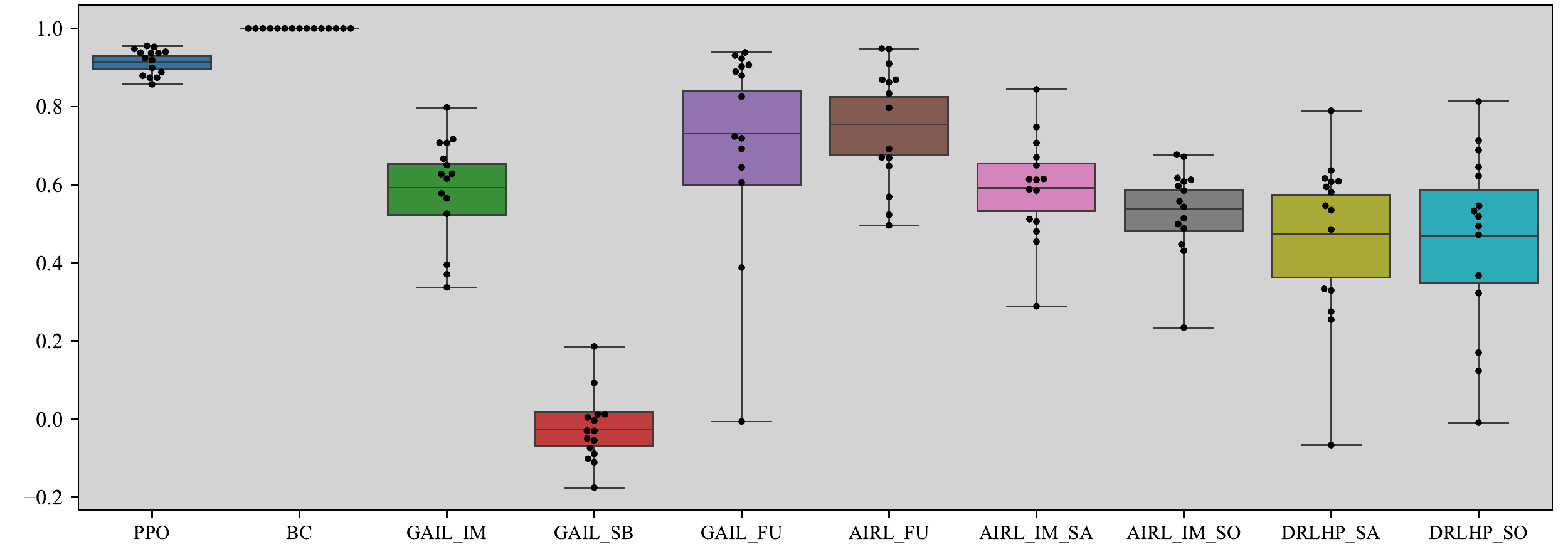}
    \caption{\procgoal{}: \bc{} achieves expert performance, while most other
      algorithms get a reasonable, but lower score, while also exhibiting high variance between seeds.
      While this task requires generalization, the fact that the states and
      actions are discrete might make it easier for \bc{} to generalize,
      compared to \parabola{} or \sort{}, where it performs poorly.
      One interesting result is \airlfu{} performing better than \airlimsa{},
      while \airlimsa{} performs better than \airlfu{} in other tasks.}
\end{figure}

\begin{figure}[h!]
    \centering
    \taskplot{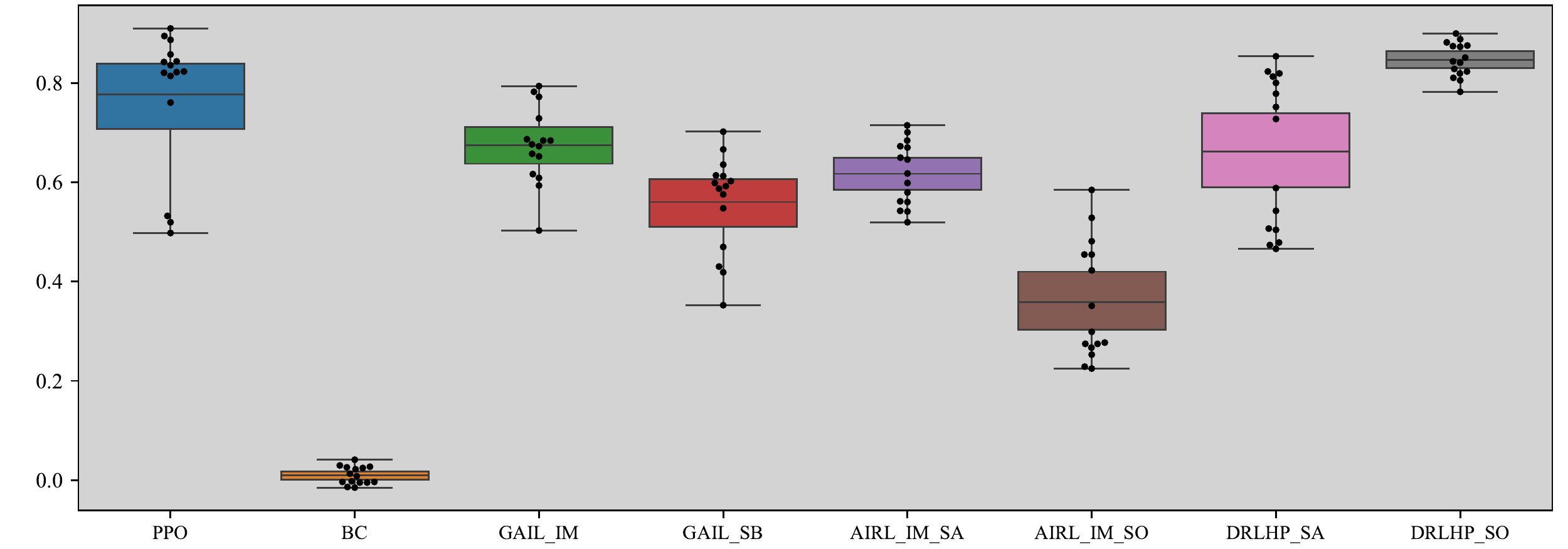}
    \caption{\sort{}: Most algorithms achieve reasonable (but sub-expert)
      performance. Intriguingly \drlhpso{} achieves higher returns than most
      algorithms, with low-variance. Learning a good policy in this task is
      challenging, given that even PPO did not get at expert performance in all
      seeds. \bc{} fails to get any reward.
      We also have that \drlhpso{} performs better than \drlhpsa{}, while
      \airlimsa{} performs better than \airlimso{}. Having a state-only reward
      might be easier to learn because there are less parameters and the
      groundtruth reward is indeed state-only, but state-action rewards can also
      incentivize the right policy by giving higher rewards to the correct
      action, making planning easier.
    }
\end{figure}

\FloatBarrier

\subsection{Algorithms}
\label{sec:appendix:algo-results}

\begin{figure}[h!]
    \centering
    \algoplot{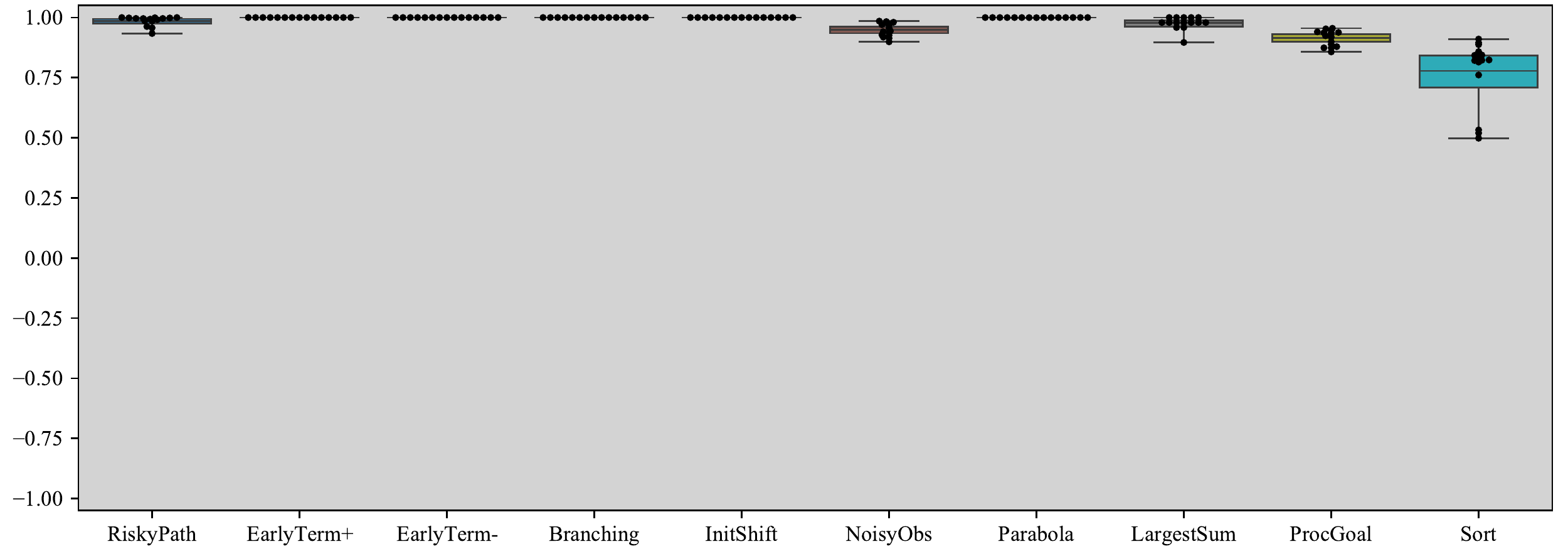}
    \caption{\ppo{}: serves as an RL baseline. We would expect most reward and imitation learning algorithms to obtain lower return, since they must learn a policy without knowing the reward. Most seeds achieve close to expert performance.}
\end{figure}

\begin{figure}[h!]
    \centering
    \algoplot{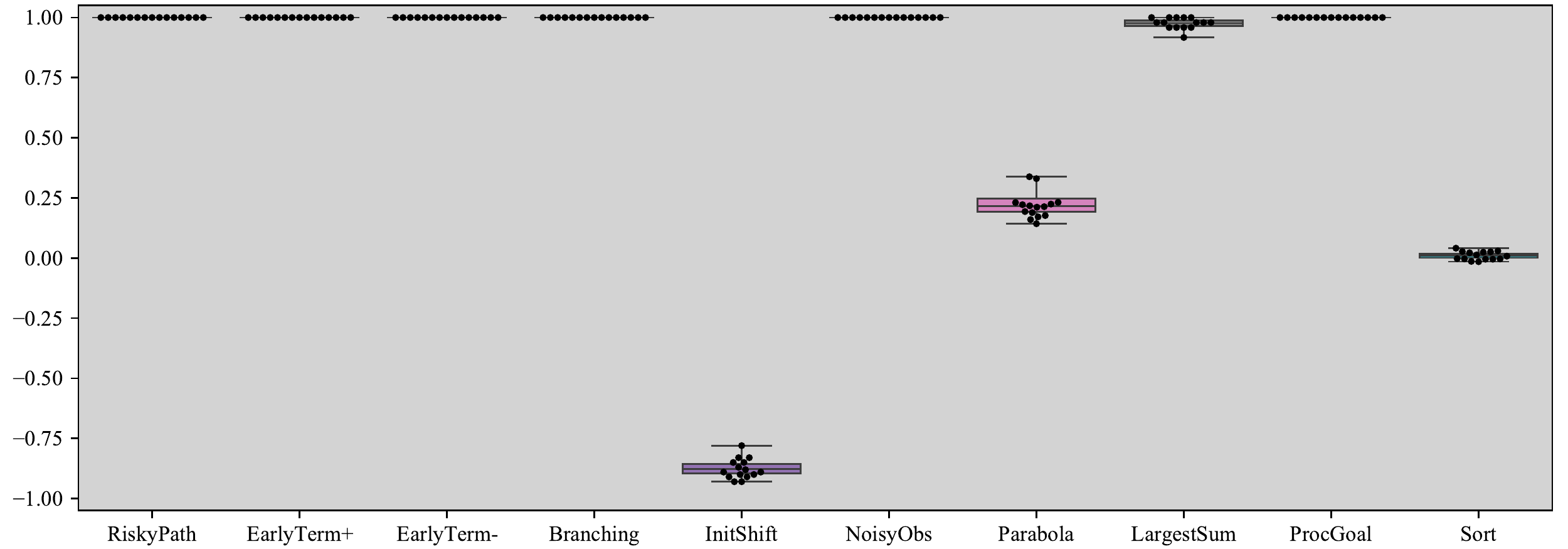}
    \caption{\bc{}: exhibits bimodal performance, either attaining near-expert return ($1.0$, normalized) in an environment or close to random ($0.0$). The return is similar across seeds. Behavioral cloning attains relatively low returns in \parabola{} and \sort{}, which have continuous observation spaces that require generalization and sequential decision making.}
\end{figure}

\begin{figure}[h!]
  \begin{subfigure}{\textwidth}
    \algoplot{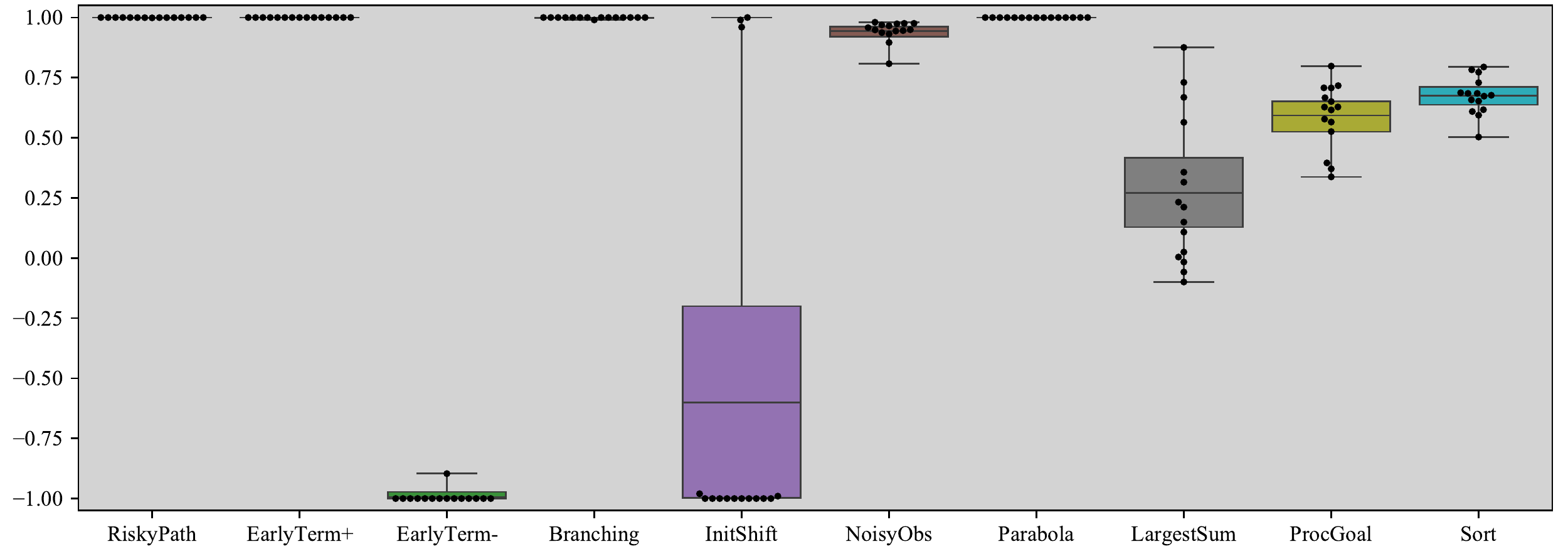}
    \caption{\gailim{}}
  \end{subfigure}
  \begin{subfigure}{\textwidth}
    \algoplot{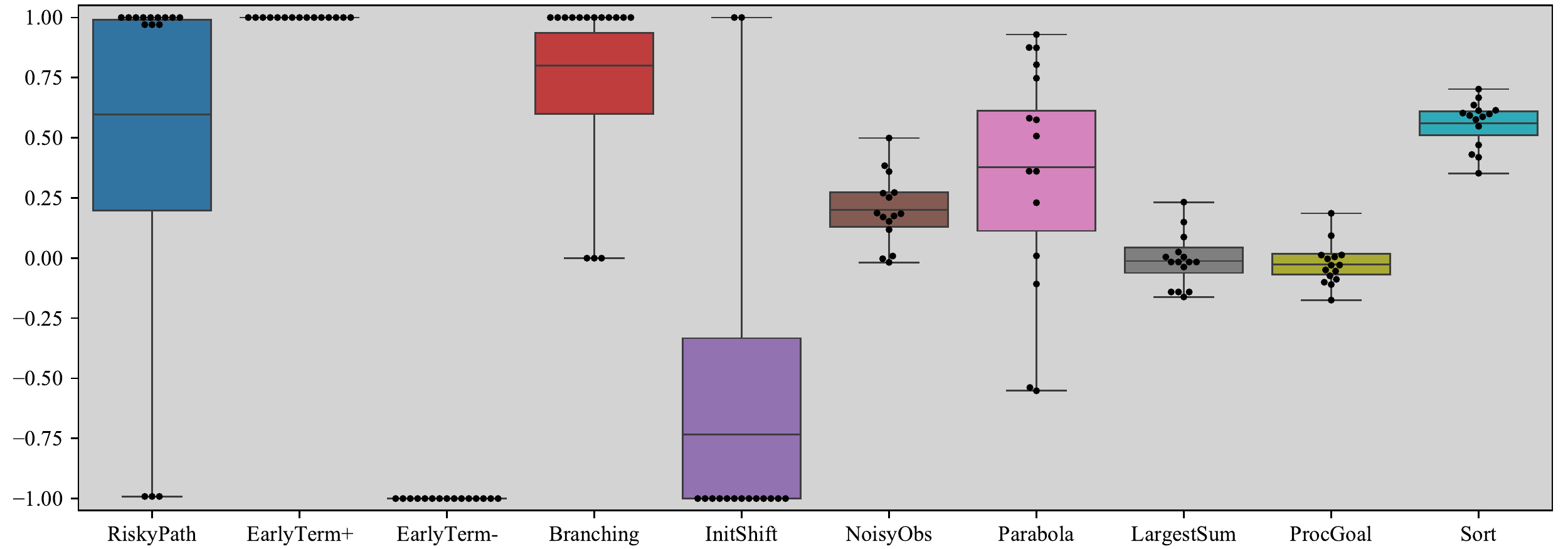}
    \caption{\gailsb{}}
  \end{subfigure}
  \begin{subfigure}{\textwidth}
    \algoplot{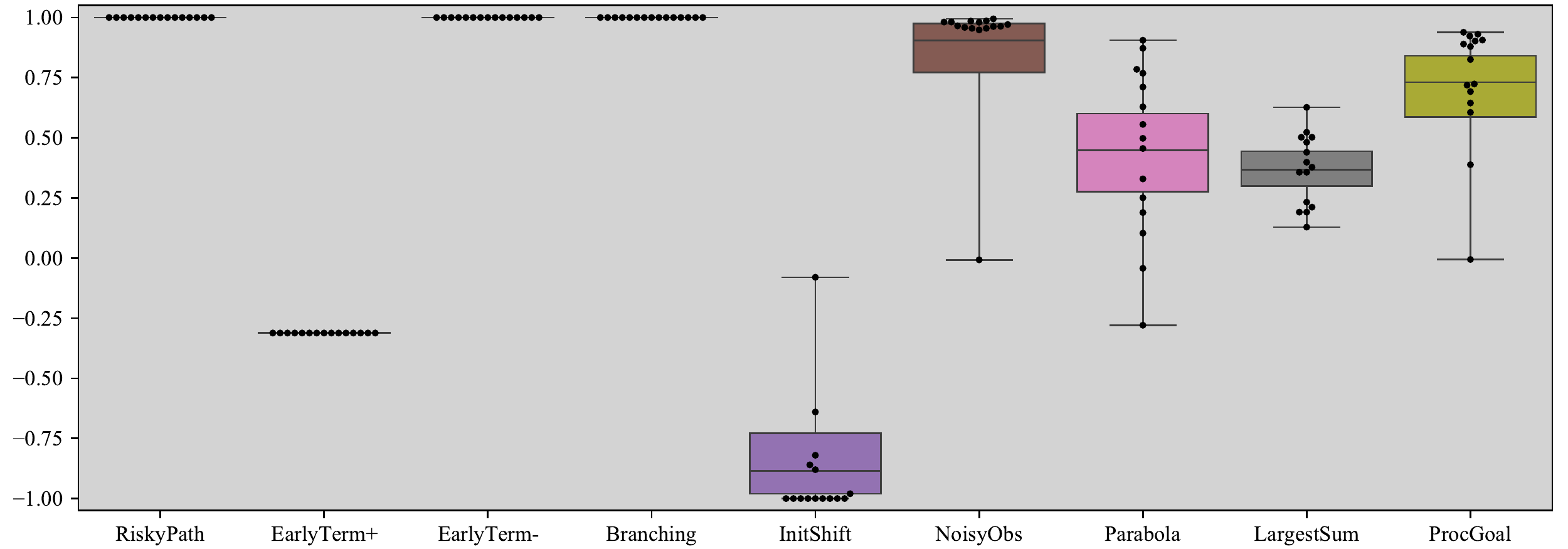}
    \caption{\gailfu{}}
  \end{subfigure}
  \caption{\gail{}: \gailim{} and \gailsb{} is positively biased, while \gailfu{} is
negatively biased, in the sense discussed in \citet{kostrikov2018discriminator}. We also see that \gailim{} results dominate \gailsb{}, with
\gailim{} performing better on every task. There are no \sort{} results for \gailfu{} because this implementation did not support the pair action space used in \sort{}.}
\end{figure}

\begin{figure}[h!]
  \begin{subfigure}{\textwidth}
    \algoplot{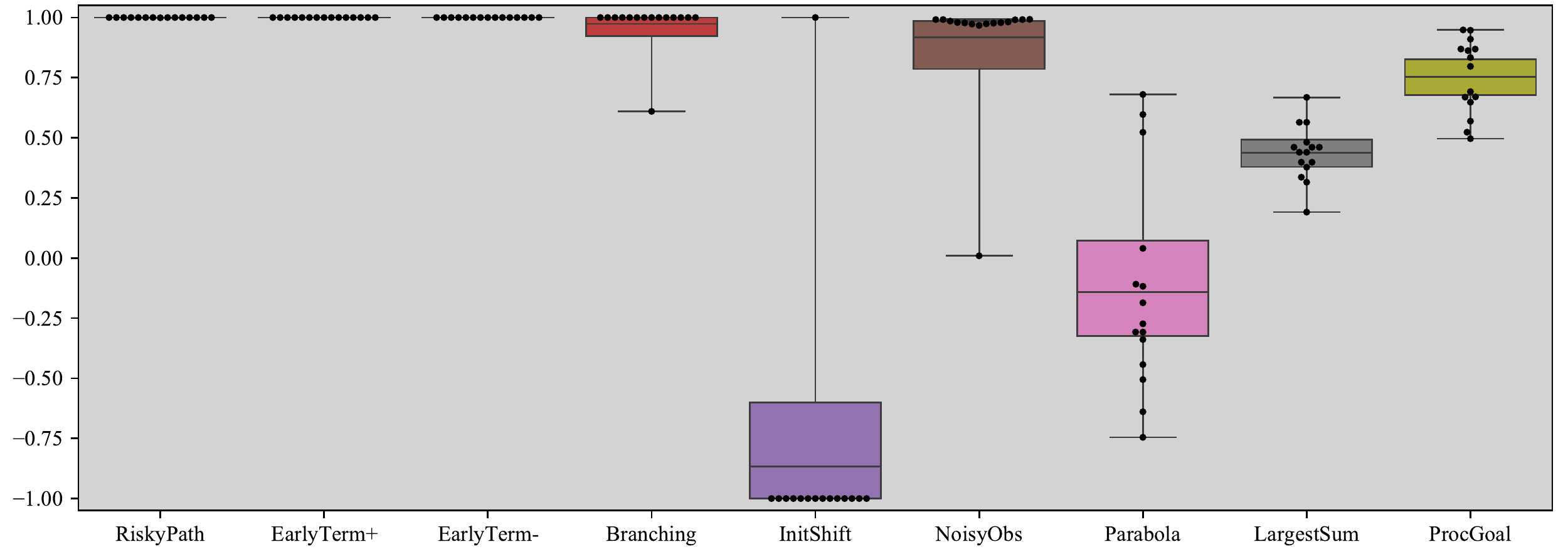}
    \caption{\airlfu{}}
  \end{subfigure}
  \begin{subfigure}{\textwidth}
    \algoplot{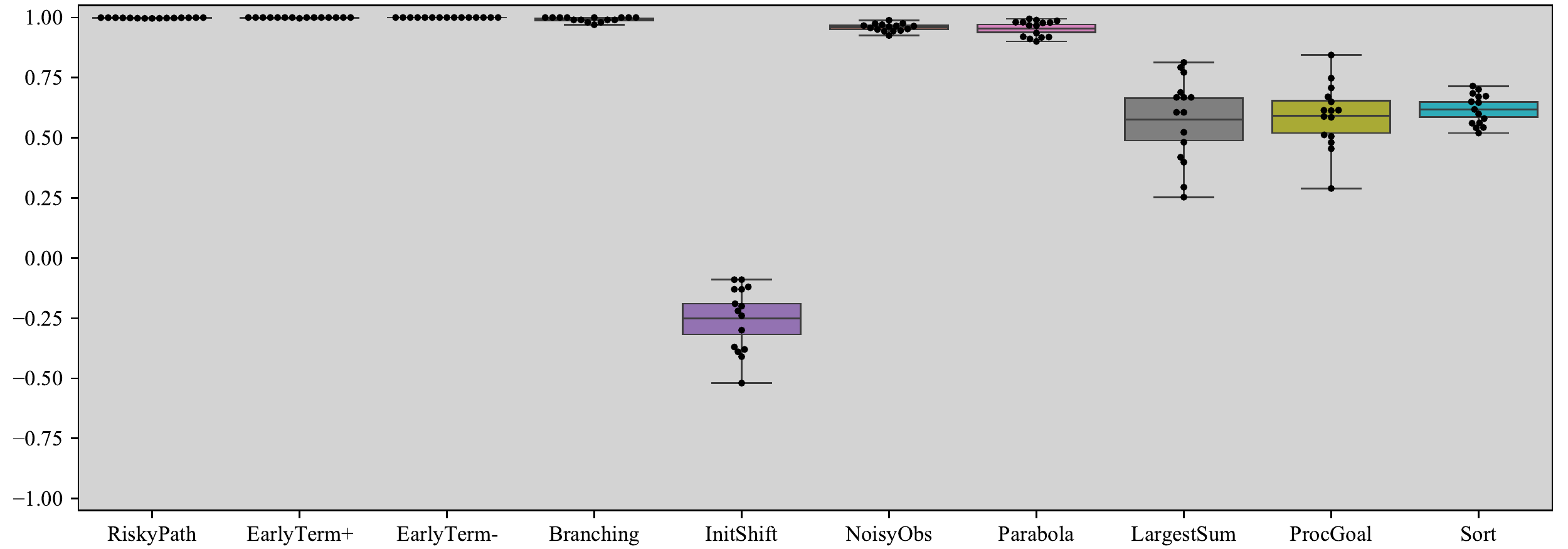}
    \caption{\airlimsa{}}
  \end{subfigure}
  \begin{subfigure}{\textwidth}
    \algoplot{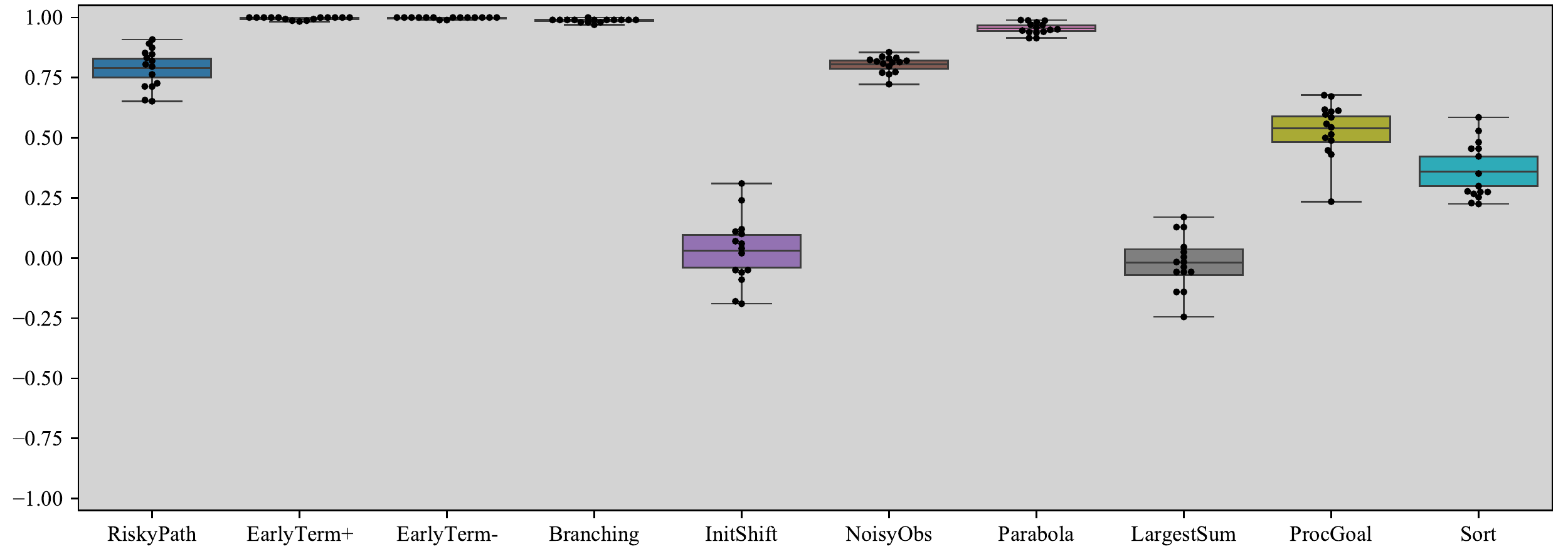}
    \caption{\airlimso{}}
  \end{subfigure}
  \caption{\airl{}: \airlimsa{} achieves higher returns than \airlimso{} and \airlfu{} on the majority of environments.
    \airlfu{} obtains particularly low returns on \parabola{}.
    \airlimsa{} performs particularly well overall, with lower variance episode return than most algorithms while attaining high return in most tasks.}
\end{figure}

\begin{figure}[h!]
  \begin{subfigure}{\textwidth}
    \algoplot{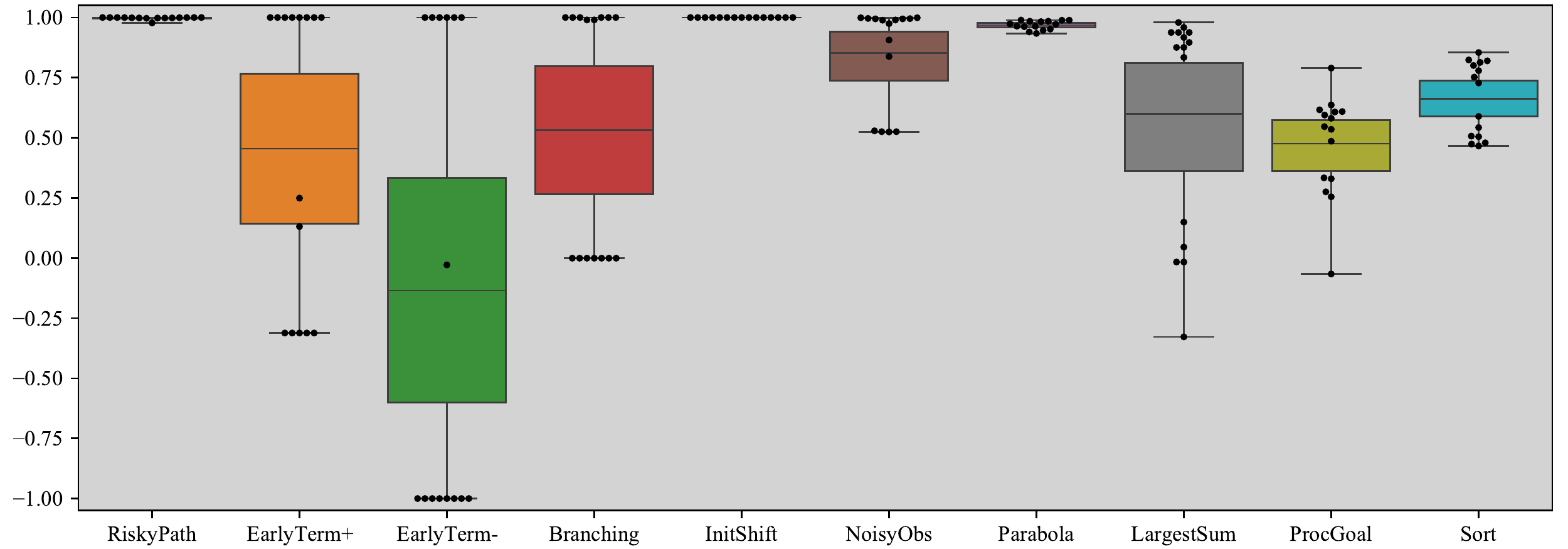}
    \caption{\drlhpsa{}}
  \end{subfigure}
  \begin{subfigure}{\textwidth}
    \algoplot{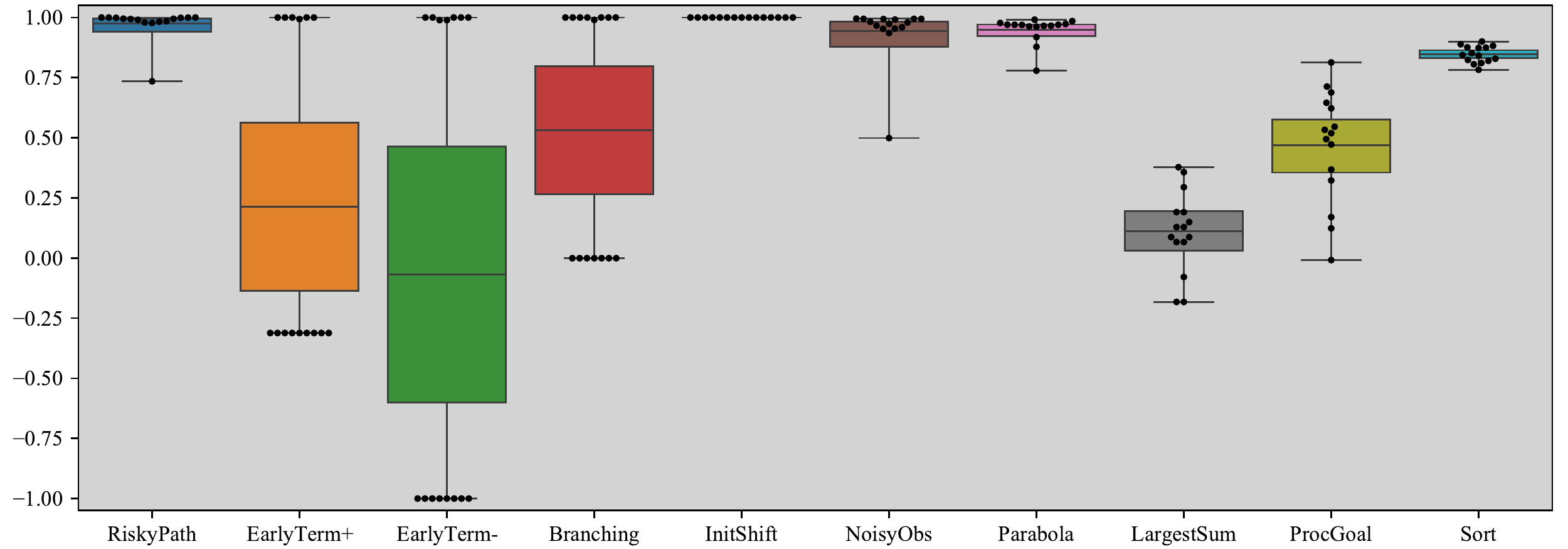}
    \caption{\drlhpso{}}
  \end{subfigure}
  \caption{\drlhp{}: achieves reasonable mean return, but exhibits high variance between seeds: some achieve expert performance while others are little better than random. Notably, \drlhpso{} is the highest scoring LfH algorithm we have tested in \sort{}. \drlhp{} performs poorly in \branch{} because of the difficulty of exploration, and in \earlypn{} because the implementation does not account for episode termination. It is the only algorithm we tested that succeeds in \initshift{}, as expected.}
\end{figure}

\begin{figure}[h!]
  \begin{subfigure}{\textwidth}
    \algoplot{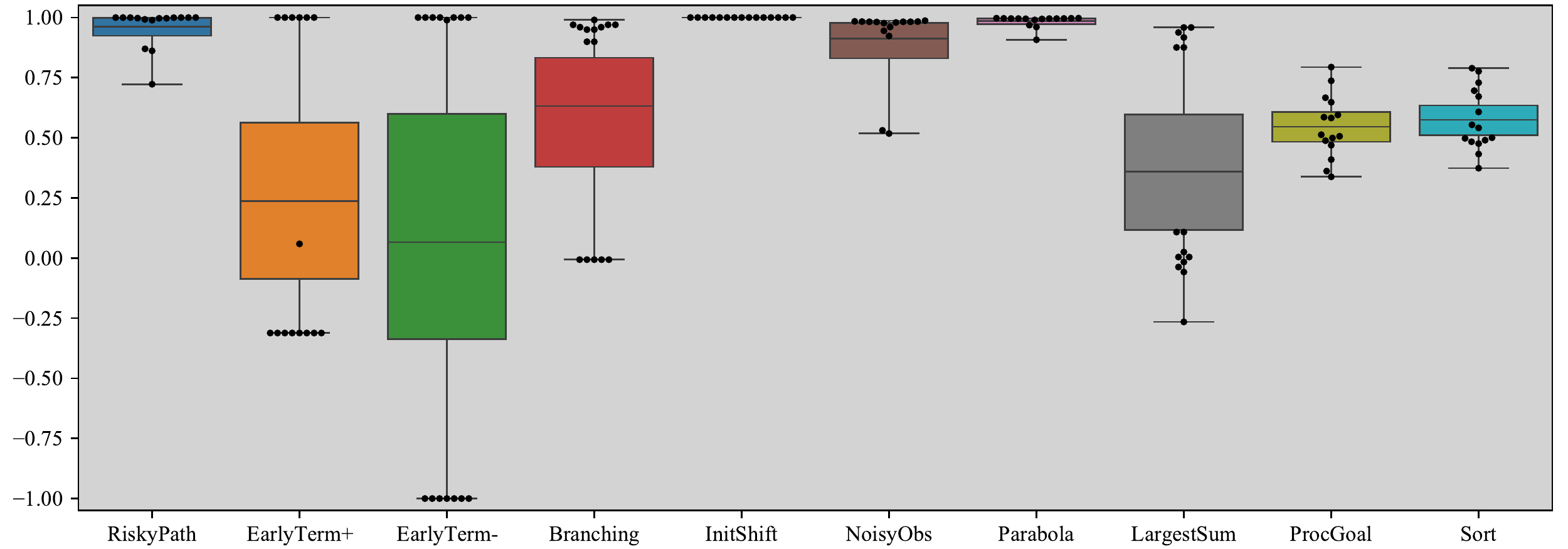}
    \caption{\drlhpa{}}
  \end{subfigure}
  \begin{subfigure}{\textwidth}
    \algoplot{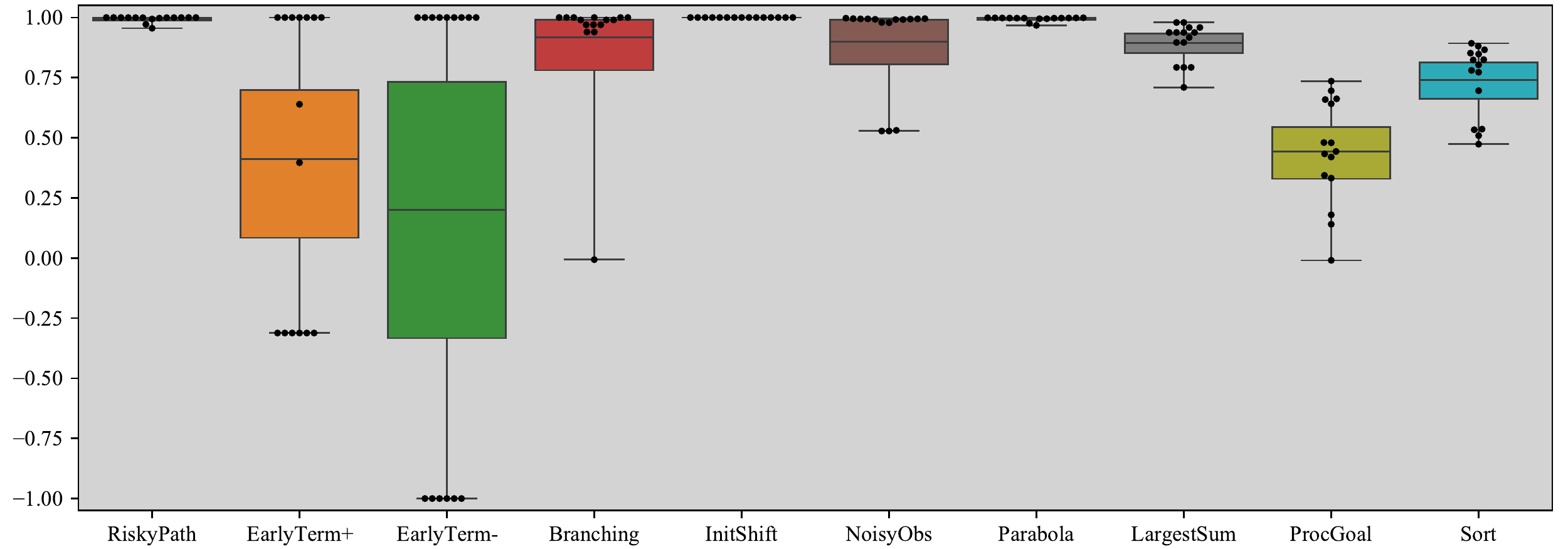}
    \caption{\drlhpb{}}
  \end{subfigure}
  \begin{subfigure}{\textwidth}
    \algoplot{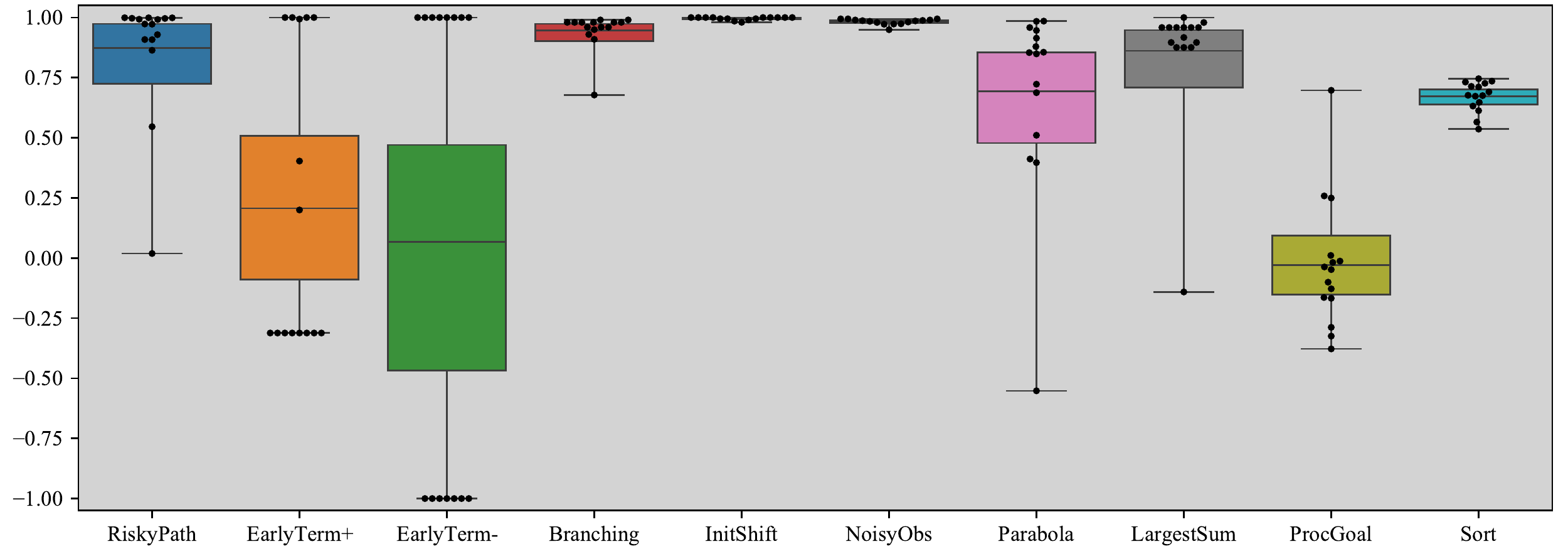}
    \caption{\drlhpc{}}
  \end{subfigure}
  \caption{\drlhp{} modifications: all versions improve the scores on \noisyobs{}; \drlhpb{} and \drlhpc{} get better scores on \branch{} and \largest{}. \drlhpb{} achives greater or similar scores to \drlhpsa{} in all tasks. Further discussion of these results can be found in Section~\ref{sec:case-study}.}
\end{figure}

\begin{figure}[h!]
  \algoplot{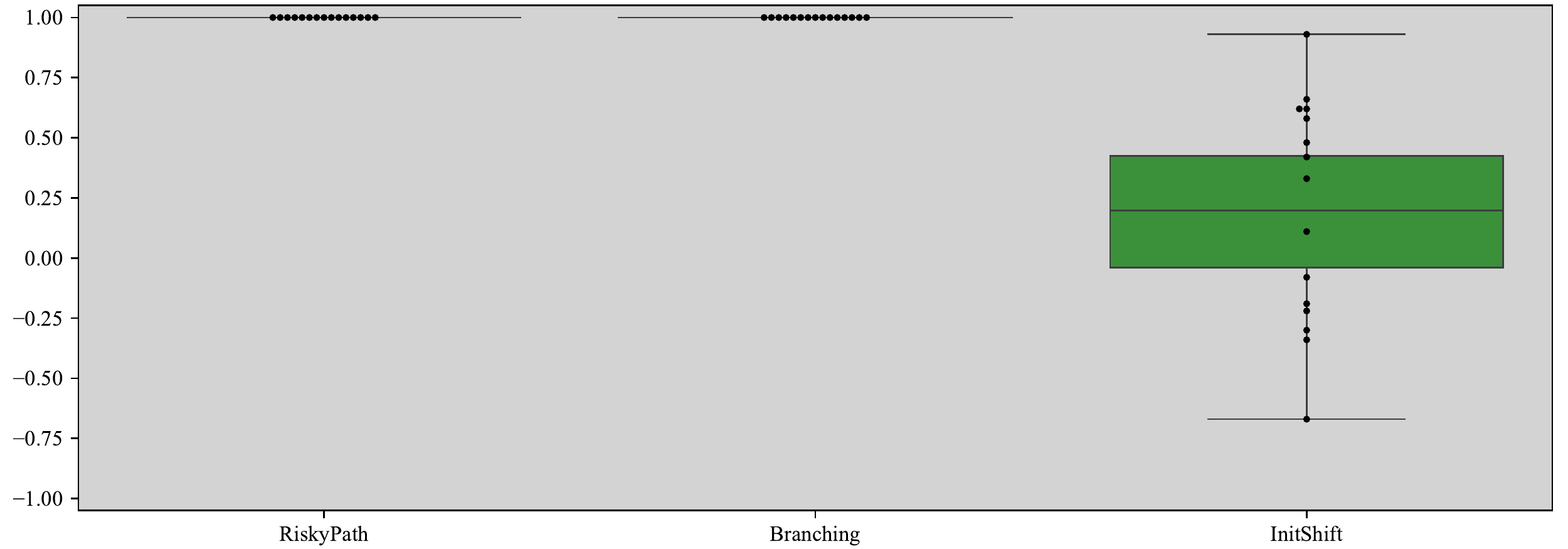}
  \algoplot{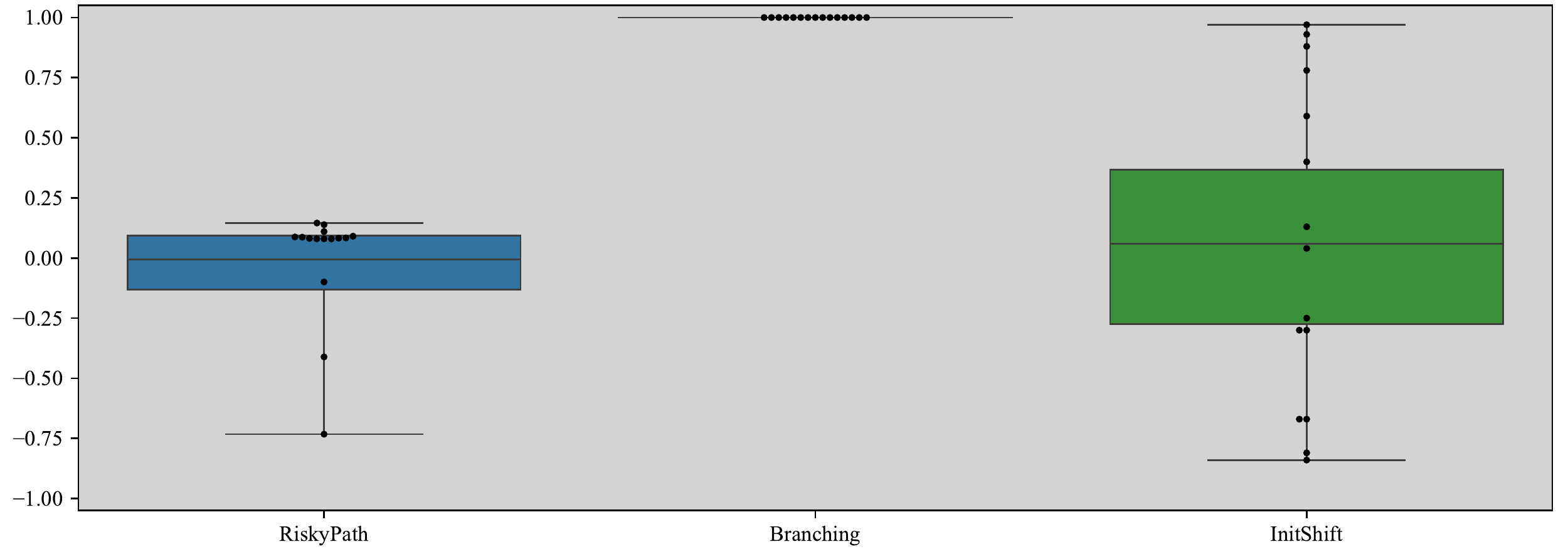}
  \caption{Tabular IRL: executed only on tasks with discrete state and action spaces and fixed horizon. As expected, \maxent{} obtains a low return in \risky{}. For \initstate{}, the expert demonstrations do not provide information to choose between the subset of states accessible during learning, and so the algorithm gets a random score that depends on the randomly initialized initial reward.}
\end{figure}

%% file: main.bbl
\begin{thebibliography}{33}
\providecommand{\natexlab}[1]{#1}
\providecommand{\url}[1]{\texttt{#1}}
\expandafter\ifx\csname urlstyle\endcsname\relax
  \providecommand{\doi}[1]{doi: #1}\else
  \providecommand{\doi}{doi: \begingroup \urlstyle{rm}\Url}\fi

\bibitem[Burda et~al.(2019)Burda, Edwards, Storkey, and
  Klimov]{burda2018exploration}
Yuri Burda, Harrison Edwards, Amos Storkey, and Oleg Klimov.
\newblock Exploration by random network distillation.
\newblock In \emph{International Conference on Learning Representations}, 2019.
\newblock URL \url{https://openreview.net/forum?id=H1lJJnR5Ym}.

\bibitem[Cabi et~al.(2019)Cabi, Colmenarejo, Novikov, Konyushkova, Reed, Jeong,
  Zolna, Aytar, Budden, Vecerik, Sushkov, Barker, Scholz, Denil, de~Freitas,
  and Wang]{cabi:2019}
Serkan Cabi, Sergio~Gómez Colmenarejo, Alexander Novikov, Ksenia Konyushkova,
  Scott Reed, Rae Jeong, Konrad Zolna, Yusuf Aytar, David Budden, Mel Vecerik,
  Oleg Sushkov, David Barker, Jonathan Scholz, Misha Denil, Nando de~Freitas,
  and Ziyu Wang.
\newblock A framework for data-driven robotics.
\newblock arXiv: 1909.12200v1 [cs.RO], 2019.

\bibitem[Christiano et~al.(2017)Christiano, Leike, Brown, Martic, Legg, and
  Amodei]{christiano:2017}
Paul~F Christiano, Jan Leike, Tom Brown, Miljan Martic, Shane Legg, and Dario
  Amodei.
\newblock Deep reinforcement learning from human preferences.
\newblock In \emph{NIPS}, pages 4299--4307, 2017.

\bibitem[Engstrom et~al.(2020)Engstrom, Ilyas, Santurkar, Tsipras, Janoos,
  Rudolph, and Madry]{engstrom:2020}
Logan Engstrom, Andrew Ilyas, Shibani Santurkar, Dimitris Tsipras, Firdaus
  Janoos, Larry Rudolph, and Aleksander Madry.
\newblock Implementation matters in deep {RL}: A case study on {PPO} and
  {TRPO}.
\newblock In \emph{ICLR}, 2020.
\newblock URL \url{https://openreview.net/forum?id=r1etN1rtPB}.

\bibitem[Fu(2018)]{fu-inverse-rl:2018}
Justin Fu.
\newblock Inverse {RL}: Implementations for imitation learning/{IRL} algorithms
  in {rllab}.
\newblock \url{https://github.com/justinjfu/inverse_rl}, 2018.

\bibitem[Fu et~al.(2018)Fu, Luo, and Levine]{fu:2018}
Justin Fu, Katie Luo, and Sergey Levine.
\newblock Learning robust rewards with adverserial inverse reinforcement
  learning.
\newblock In \emph{ICLR}, 2018.

\bibitem[Gleave(2020)]{evaluating-rewards:2020}
Adam Gleave.
\newblock Evaluating rewards: comparing and evaluating reward models.
\newblock \url{https://github.com/humancompatibleai/evaluating-rewards}, 2020.

\bibitem[Guyon and Elisseeff(2003)]{guyon:2003}
Isabelle Guyon and André Elisseeff.
\newblock An introduction to variable and feature selection.
\newblock \emph{JMLR}, pages 1157--1182, 3 2003.

\bibitem[Henderson et~al.(2018)Henderson, Islam, Bachman, Pineau, Precup, and
  Meger]{henderson2018deep}
Peter Henderson, Riashat Islam, Philip Bachman, Joelle Pineau, Doina Precup,
  and David Meger.
\newblock Deep reinforcement learning that matters.
\newblock In \emph{AAAI}, 2018.

\bibitem[Hill et~al.(2018)Hill, Raffin, Ernestus, Gleave, Kanervisto, Traore,
  Dhariwal, Hesse, Klimov, Nichol, Plappert, Radford, Schulman, Sidor, and
  Wu]{stable-baselines:2018}
Ashley Hill, Antonin Raffin, Maximilian Ernestus, Adam Gleave, Anssi
  Kanervisto, Rene Traore, Prafulla Dhariwal, Christopher Hesse, Oleg Klimov,
  Alex Nichol, Matthias Plappert, Alec Radford, John Schulman, Szymon Sidor,
  and Yuhuai Wu.
\newblock {Stable Baselines}.
\newblock \url{https://github.com/hill-a/stable-baselines}, 2018.

\bibitem[Ho and Ermon(2016)]{ho:2016}
Jonathan Ho and Stefano Ermon.
\newblock Generative adversarial imitation learning.
\newblock In \emph{NIPS}, pages 4565--4573, 2016.

\bibitem[Islam et~al.(2017)Islam, Henderson, Gomrokchi, and
  Precup]{islam2017reproducibility}
Riashat Islam, Peter Henderson, Maziar Gomrokchi, and Doina Precup.
\newblock Reproducibility of benchmarked deep reinforcement learning tasks for
  continuous control.
\newblock \emph{arXiv preprint arXiv:1708.04133}, 2017.

\bibitem[James et~al.(2020)James, Ma, Arrojo, and Davison]{james:2020}
Stephen James, Zicong Ma, David~Rovick Arrojo, and Andrew~J. Davison.
\newblock Rlbench: The robot learning benchmark learning environment.
\newblock \emph{IEEE Robotics and Automation Letters}, 5\penalty0 (2):\penalty0
  3019--3026, 2020.

\bibitem[Johnson et~al.(2017)Johnson, Hariharan, van~der Maaten, Fei-Fei,
  Lawrence~Zitnick, and Girshick]{johnson:2017}
Justin Johnson, Bharath Hariharan, Laurens van~der Maaten, Li~Fei-Fei,
  C.~Lawrence~Zitnick, and Ross Girshick.
\newblock {CLEVR}: A diagnostic dataset for compositional language and
  elementary visual reasoning.
\newblock In \emph{CVPR}, 2017.

\bibitem[Kostrikov et~al.(2018)Kostrikov, Agrawal, Dwibedi, Levine, and
  Tompson]{kostrikov2018discriminator}
Ilya Kostrikov, Kumar~Krishna Agrawal, Debidatta Dwibedi, Sergey Levine, and
  Jonathan Tompson.
\newblock Discriminator-actor-critic: Addressing sample inefficiency and reward
  bias in adversarial imitation learning.
\newblock \emph{arXiv preprint arXiv:1809.02925}, 2018.

\bibitem[Kottur et~al.(2019)Kottur, Moura, Parikh, Batra, and
  Rohrbach]{kottur:2019}
Satwik Kottur, Jos{\'{e}} M.~F. Moura, Devi Parikh, Dhruv Batra, and Marcus
  Rohrbach.
\newblock {CLEVR}-{D}ialog: {A} diagnostic dataset for multi-round reasoning in
  visual dialog.
\newblock In \emph{NAACL-HLT}, 2019.

\bibitem[Liu et~al.(2019)Liu, Liu, Bai, and Yuille]{liu:2019}
Runtao Liu, Chenxi Liu, Yutong Bai, and Alan~L. Yuille.
\newblock {CLEVR}-{R}ef+: Diagnosing visual reasoning with referring
  expressions.
\newblock In \emph{CVPR}, 2019.

\bibitem[Memmesheimer et~al.(2019)Memmesheimer, Kramer, Seib, and
  Paulus]{memmesheimer:2019}
Raphael Memmesheimer, Ivanna Kramer, Viktor Seib, and Dietrich Paulus.
\newblock Simitate: A hybrid imitation learning benchmark.
\newblock In \emph{IROS}, pages 5243--5249, 2019.

\bibitem[Myers et~al.(2011)Myers, Sandler, and Badgett]{myers2011art}
Glenford~J Myers, Corey Sandler, and Tom Badgett.
\newblock \emph{The art of software testing}, chapter~5.
\newblock John Wiley \& Sons, 2011.

\bibitem[Ng and Russell(2000)]{ng:2000}
Andrew~Y. Ng and Stuart Russell.
\newblock Algorithms for inverse reinforcement learning.
\newblock In \emph{ICML}, 2000.

\bibitem[OpenAI et~al.(2019)OpenAI, Akkaya, Andrychowicz, Chociej, Litwin,
  McGrew, Petron, Paino, Plappert, Powell, Ribas, Schneider, Tezak, Tworek,
  Welinder, Weng, Yuan, Zaremba, and Zhang]{openai:2019}
OpenAI, Ilge Akkaya, Marcin Andrychowicz, Maciek Chociej, Mateusz Litwin, Bob
  McGrew, Arthur Petron, Alex Paino, Matthias Plappert, Glenn Powell, Raphael
  Ribas, Jonas Schneider, Nikolas Tezak, Jerry Tworek, Peter Welinder, Lilian
  Weng, Qiming Yuan, Wojciech Zaremba, and Lei Zhang.
\newblock Solving {Rubik's Cube} with a robot hand.
\newblock arXiv: 1910.07113v1 [cs.LG], 2019.

\bibitem[Osband et~al.(2020)Osband, Doron, Hessel, Aslanides, Sezener, Saraiva,
  McKinney, Lattimore, Szepesvari, Singh, Roy, Sutton, Silver, and
  Hasselt]{osband:2020}
Ian Osband, Yotam Doron, Matteo Hessel, John Aslanides, Eren Sezener, Andre
  Saraiva, Katrina McKinney, Tor Lattimore, Csaba Szepesvari, Satinder Singh,
  Benjamin~Van Roy, Richard Sutton, David Silver, and Hado~Van Hasselt.
\newblock Behaviour suite for reinforcement learning.
\newblock In \emph{ICLR}, 2020.

\bibitem[Reddy et~al.(2020)Reddy, Dragan, and Levine]{reddy:2020}
Siddharth Reddy, Anca~D. Dragan, and Sergey Levine.
\newblock {SQIL}: Imitation learning via reinforcement learning with sparse
  rewards.
\newblock In \emph{ICLR}, 2020.

\bibitem[Ross et~al.(2011)Ross, Gordon, and Bagnell]{ross:2011}
Stephane Ross, Geoffrey Gordon, and Drew Bagnell.
\newblock A reduction of imitation learning and structured prediction to
  no-regret online learning.
\newblock In \emph{AISTATS}, 2011.

\bibitem[Schulman et~al.(2017)Schulman, Wolski, Dhariwal, Radford, and
  Klimov]{schulman:2017}
John Schulman, Filip Wolski, Prafulla Dhariwal, Alec Radford, and Oleg Klimov.
\newblock Proximal policy optimization algorithms.
\newblock arXiv:1707.06347v2 [cs.LG], 2017.

\bibitem[Silver et~al.(2016)Silver, Huang, Maddison, Guez, Sifre, van~den
  Driessche, Schrittwieser, Antonoglou, Panneershelvam, Lanctot, Dieleman,
  Grewe, Nham, Kalchbrenner, Sutskever, Lillicrap, Leach, Kavukcuoglu, Graepel,
  and Hassabis]{silver:2016}
David Silver, Aja Huang, Chris~J. Maddison, Arthur Guez, Laurent Sifre, George
  van~den Driessche, Julian Schrittwieser, Ioannis Antonoglou, Veda
  Panneershelvam, Marc Lanctot, Sander Dieleman, Dominik Grewe, John Nham, Nal
  Kalchbrenner, Ilya Sutskever, Timothy Lillicrap, Madeleine Leach, Koray
  Kavukcuoglu, Thore Graepel, and Demis Hassabis.
\newblock Mastering the game of {Go} with deep neural networks and tree search.
\newblock \emph{Nature}, 529\penalty0 (7587):\penalty0 484--489, 2016.

\bibitem[Sinha et~al.(2019)Sinha, Sodhani, Dong, Pineau, and
  Hamilton]{sinha:2019}
Koustuv Sinha, Shagun Sodhani, Jin Dong, Joelle Pineau, and William~L.
  Hamilton.
\newblock {CLUTRR:} {A} diagnostic benchmark for inductive reasoning from text.
\newblock In \emph{EMNLP}, 2019.

\bibitem[Vinyals et~al.(2019)Vinyals, Babuschkin, Czarnecki, Mathieu, Dudzik,
  Chung, Choi, Powell, Ewalds, Georgiev, Oh, Horgan, Kroiss, Danihelka, Huang,
  Sifre, Cai, Agapiou, Jaderberg, Vezhnevets, Leblond, Pohlen, Dalibard,
  Budden, Sulsky, Molloy, Paine, Gulcehre, Wang, Pfaff, Wu, Ring, Yogatama,
  W{\"u}nsch, McKinney, Smith, Schaul, Lillicrap, Kavukcuoglu, Hassabis, Apps,
  and Silver]{vinyals:2019}
Oriol Vinyals, Igor Babuschkin, Wojciech~M. Czarnecki, Micha{\"e}l Mathieu,
  Andrew Dudzik, Junyoung Chung, David~H. Choi, Richard Powell, Timo Ewalds,
  Petko Georgiev, Junhyuk Oh, Dan Horgan, Manuel Kroiss, Ivo Danihelka, Aja
  Huang, Laurent Sifre, Trevor Cai, John~P. Agapiou, Max Jaderberg,
  Alexander~S. Vezhnevets, R{\'e}mi Leblond, Tobias Pohlen, Valentin Dalibard,
  David Budden, Yury Sulsky, James Molloy, Tom~L. Paine, Caglar Gulcehre, Ziyu
  Wang, Tobias Pfaff, Yuhuai Wu, Roman Ring, Dani Yogatama, Dario W{\"u}nsch,
  Katrina McKinney, Oliver Smith, Tom Schaul, Timothy Lillicrap, Koray
  Kavukcuoglu, Demis Hassabis, Chris Apps, and David Silver.
\newblock Grandmaster level in {StarCraft II} using multi-agent reinforcement
  learning.
\newblock \emph{Nature}, 575\penalty0 (7782):\penalty0 350--354, 2019.

\bibitem[Wacker(2015)]{wacker:2015}
Mike Wacker.
\newblock Just say no to more end-to-end tests.
\newblock Google Testing Blog, 2015.
\newblock URL
  \url{https://testing.googleblog.com/2015/04/just-say-no-to-more-end-to-end-tests.html}.

\bibitem[Wang et~al.(2019)Wang, Singh, Michael, Hill, Levy, and
  Bowman]{wang:2019}
Alex Wang, Amanpreet Singh, Julian Michael, Felix Hill, Omer Levy, and
  Samuel~R. Bowman.
\newblock {GLUE}: A multi-task benchmark and analysis platform for natural
  language understanding.
\newblock In \emph{ICLR}, 2019.

\bibitem[Wang et~al.(2020)Wang, Gleave, and Toyer]{imitation:2020}
Steven Wang, Adam Gleave, and Sam Toyer.
\newblock imitation: implementations of inverse reinforcement learning and
  imitation learning algorithms.
\newblock \url{https://github.com/humancompatibleai/imitation}, 2020.

\bibitem[Ziebart(2010)]{ziebart:2010:thesis}
Brian~D Ziebart.
\newblock \emph{Modeling purposeful adaptive behavior with the principle of
  maximum causal entropy}.
\newblock PhD thesis, Carnegie Mellon University, 2010.

\bibitem[Ziebart et~al.(2008)Ziebart, Maas, Bagnell, and Dey]{ziebart:2008}
Brian~D. Ziebart, Andrew Maas, J.~Andrew Bagnell, and Anind~K. Dey.
\newblock Maximum entropy inverse reinforcement learning.
\newblock In \emph{AAAI}, 2008.

\end{thebibliography}
